\documentclass[]{alibaba-nlp}
\usepackage{hyperref}
\usepackage{url}
\usepackage{amsmath, amsthm}
\usepackage{enumitem}
\usepackage{booktabs}
\usepackage{multirow}
\usepackage{multicol}
\usepackage{makecell}
\usepackage{graphicx}
\usepackage{amssymb}
\usepackage{wrapfig}
\usepackage{threeparttable}
\usepackage{tabularx}
\usepackage[table]{xcolor}
\usepackage{microtype}
\usepackage{titletoc}
\usepackage{subcaption}
\usepackage{adjustbox}

\usepackage{listings,xcolor,fontawesome5}
\usepackage[most]{tcolorbox}

\definecolor{prompt-bg}{RGB}{250,250,255}
\definecolor{prompt-border}{RGB}{70,90,140}
\definecolor{prompt-title-bg}{RGB}{70,90,140}
\definecolor{prompt-title-fg}{RGB}{255,255,255}
\definecolor{prompt-leftbar}{RGB}{90,110,160}

\lstdefinestyle{promptstyle}{
    basicstyle=\ttfamily\footnotesize,
    breaklines=true,
    backgroundcolor=\color{prompt-bg},
    frame=none, numbers=none,
    columns=fullflexible,
    keepspaces=true,
    showspaces=false,
    showstringspaces=false,
    tabsize=2,
    escapeinside=||,
    lineskip=-0.5pt,
}

\newtcolorbox{promptbox}[1]{
    enhanced, breakable,
    arc=2pt, boxrule=0.6pt,
    colframe=prompt-leftbar!50!prompt-border,
    colback=prompt-bg,
    coltitle=prompt-title-fg,
    fonttitle=\bfseries\footnotesize,
    title={\strut #1},
    titlerule=0pt,
    toptitle=2pt, bottomtitle=1pt,
    top=3pt, bottom=3pt, left=6pt, right=5pt,
    boxsep=0pt,
    colbacktitle=prompt-title-bg,
    before title={\faRobot\hspace{0.25em}},
}

\newtheorem{definition}{Definition}

\title{Towards Universal Video Retrieval: Generalizing Video Embedding via Synthesized Multimodal Pyramid Curriculum}

\author[1,2]{Zhuoning Guo}
\author[2]{Mingxin Li}
\author[2]{Yanzhao Zhang}
\author[2]{Dingkun Long}
\author[2]{Pengjun Xie}
\author[1]{Xiaowen Chu}

\affiliation[1]{AI Thrust, HKUST(GZ)}
\affiliation[2]{Tongyi Lab, Alibaba Group}

\abstract{
The prevailing video retrieval paradigm is structurally misaligned, as narrow benchmarks incentivize correspondingly limited data and single-task training. Therefore, universal capability is suppressed due to the absence of a diagnostic evaluation that defines and demands multi-dimensional generalization. To break this cycle, we introduce a framework built on the co-design of evaluation, data, and modeling. First, we establish the Universal Video Retrieval Benchmark (UVRB), a suite of 16 datasets designed not only to measure performance but also to diagnose critical capability gaps across tasks and domains. Second, guided by UVRB's diagnostics, we introduce a scalable synthesis workflow that generates 1.55 million high-quality pairs to populate the semantic space required for universality. Finally, we devise the Modality Pyramid, a curriculum that trains our General Video Embedder (GVE) by explicitly leveraging the latent interconnections within our diverse data. Extensive experiments show GVE achieves state-of-the-art zero-shot generalization on UVRB. In particular, our analysis reveals that popular benchmarks are poor predictors of general ability and that partially relevant retrieval is a dominant but overlooked scenario. Overall, our co-designed framework provides a practical path to escape the limited scope and advance toward truly universal video retrieval.
}

\date{\today}
\contact{Zhuoning Guo (\email{zguo772@connect.hkust-gz.edu.cn})}
\projectleader{Dingkun Long (\email{dingkun.ldk@alibaba-inc.com})}
\correspondingauthor{Xiaowen Chu (\email{xwchu@hkust-gz.edu.cn})}
\projectpage{\url{https://gzn00417.github.io/GVE/}}

\begin{document}

\maketitle

\begin{figure}[h]
    \centering
    \includegraphics[width=\linewidth]{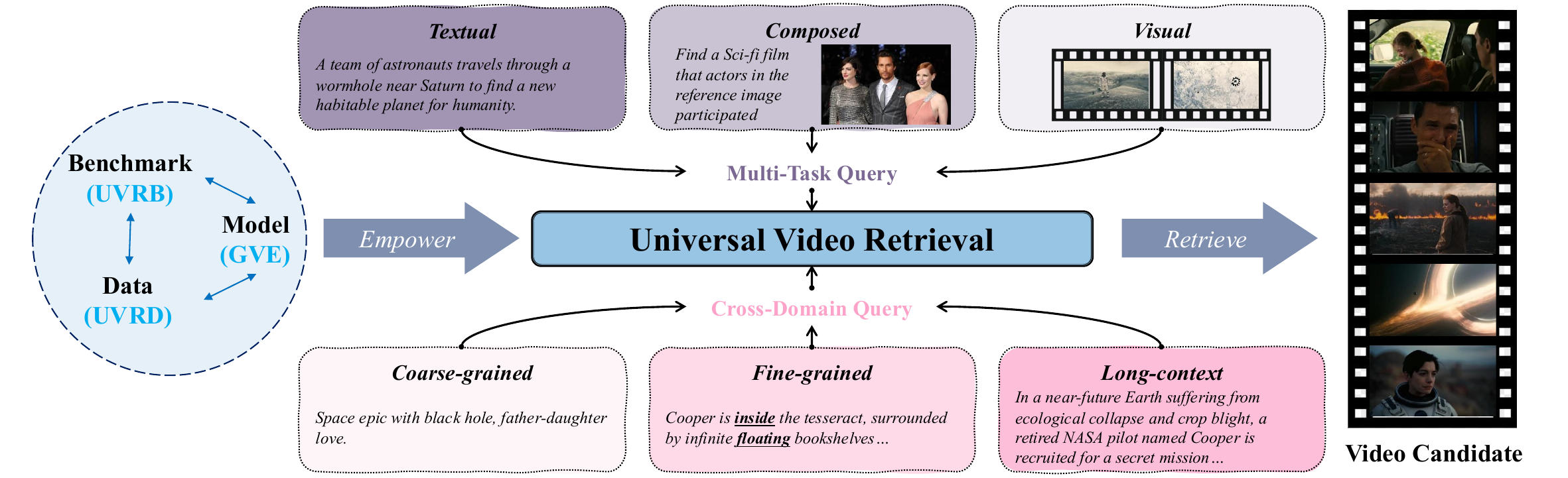}
    \caption{We propose Universal Video Retrieval~(UVR) that retrieves videos with multi-task, cross-domain queries, which can be achieved via benchmark-data-model co-design in this work.}
    \label{fig:uvr}
\end{figure}

\section{Introduction}

Video retrieval is a critical, yet challenging task for modern search engines and recommendation systems, requiring effective video embedding models~\cite{zhu2023deep}. Early efforts extended Contrastive Language-Image Pretraining (CLIP)~\cite{radford2021learning} to video~\cite{ma2022x}. Now, a paradigm shift is underway, with Multimodal Large Language Models (MLLMs) rapidly displacing CLIP for their superior language understanding and visual generalization capabilities~\cite{kong2025modality}.
Current practice involves training these models on massive datasets with simple and noisy text annotations (e.g., WebVid~\cite{bain2021frozen}) with strong results for coarse-grained text-to-video retrieval on benchmarks (e.g., MSRVTT~\cite{xu2016msr}).

However, they struggle with the complexity of diverse video retrieval scenarios~(Figure~\ref{fig:uvr}). First, a narrow semantic distribution renders these models ineffective in fine-grained queries required to understand spatial relations or temporal dynamics~\cite{xu2025carebench}, as well as in long-context retrieval within lengthy videos~\cite{cai2025lovr}. Second, the scope of applicable tasks is restricted, with little support for diverse query formats beyond plain text, such as composed retrieval using text-and-image pairs and purely visual queries. Existing specialized models~\cite {hummel2024egocvr} are costly and hinder progress toward a single, generalizable model across these emerging scenarios.


\begin{figure}[t]
    \centering
    \includegraphics[width=\linewidth]{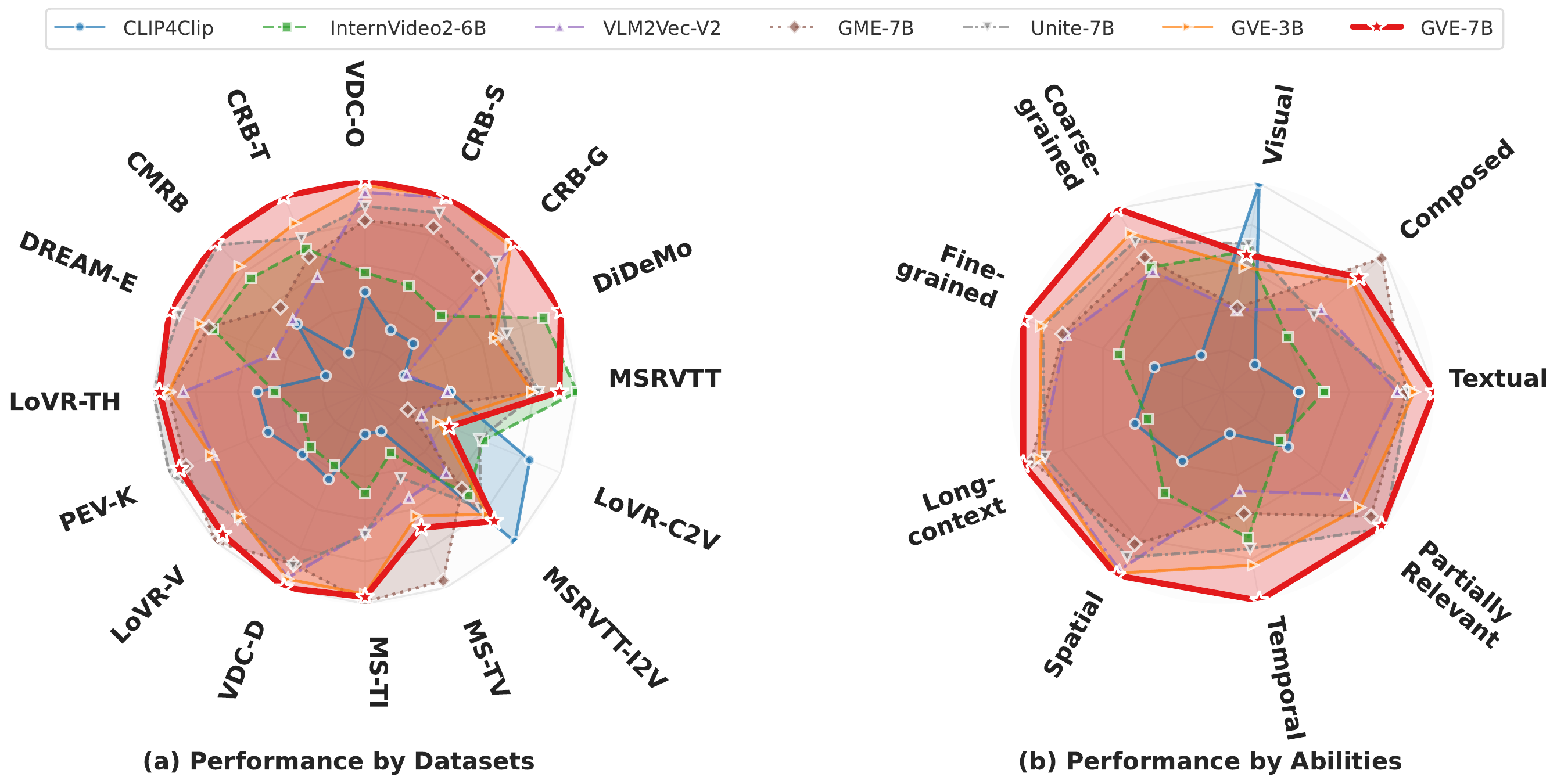}
    \caption{Model performance on UVRB for 16 datasets and 9 abilities~(3 main tasks and 6 (sub-) domains).}
    \label{fig:radar_uvrb}
\end{figure}

Therefore, to establish a framework for universal video retrieval that supports multi-domain, multi-granularity, and multi-task capabilities, three coupled challenges in evaluation, data, and modeling need to be addressed simultaneously:
(1)~\textit{Dimensional Diagnostic Evaluation}: The foundational step is to define and measure universality quantitatively. This necessitates a comprehensive evaluation framework capable not only of assessing performance across diverse tasks but, more critically, of diagnosing the intricate correlations and interferences between them.
(2)~\textit{Large-Scale Quality-Controlled Data Synthesis}: Existing datasets are either too small or are biased, while collecting a new, massive dataset is prohibitively expensive. Growing works attempt to synthesize data to address this challenge~\cite{chen2023vast,chai2024auroracap,ventura2024covr}. However, these resources often exhibit uneven quality and distribution. Therefore, gaining precise control over the properties of large-scale, cross-domain, multi-task data via a unified synthesis process is the second challenge.
(3)~\textit{Interconnected Multi-task Representation Learning}: A critical, often-ignored aspect is the inherent hierarchy among tasks. Foundational abilities such as spatial perception (e.g., object recognition) serve as building blocks for higher-order temporal reasoning (e.g., action recognition). This principle is evidenced by the remarkable success of image-only trained models like GME~\cite{Zhang_2025_CVPR} on video retrieval tasks (see Figure~\ref{fig:radar_uvrb}), highlighting latent cross-task adaptability. Nevertheless, conventional models that assume task independence fail to capitalize on this structure, which is a key to unlocking superior generalization.

To this end, we propose a holistic framework by co-designing evaluation, data, and modeling.
Specifically, we first construct the \textbf{Universal Video Retrieval Benchmark~(UVRB)}, a comprehensive suite with $16$ test datasets across diverse domains and tasks for $14$ state-of-the-art models. More importantly, an in-depth analysis quantitatively exposes the limitations of current approaches.
Second, based on the diagnostics, we design \textbf{V-SynFlow}, a multi-stage data synthesis workflow that transforms massive, low-quality text-video pairs into a high-quality and multi-task dataset, \textbf{Universal Video Retrieval Dataset~(UVRD)}. It consists of over $1.55$ million video retrieval pairs with rich spatial-temporal details, diverse descriptive styles, and distinct task formats.
Third, we devise the \textsc{Modality Pyramid}, a customized curriculum that leverages inherent task and domain dependencies to optimize a \textbf{General Video Embedder (GVE)} on the diverse synthesized data for advanced zero-shot task and domain adaptation. The bottom-up, pyramid-shaped curriculum prioritizes data-abundant, foundational tasks before progressing to more complex, dependent ones for progressive and stable knowledge acquisition.
Extensive experiments on UVRB validate the effectiveness of GVE~(Figure~\ref{fig:radar_uvrb}). Besides, the diagnostic analysis reveals underexplored findings. For example, conventional benchmarks are not representative of the overall retrieval ability, indicating the potential for the overfitting of existing models on in-domain data. Instead, partially relevant video retrieval, despite low research attention, is a typical and generalizable scenario in this field.

The contributions of this work are summarized as follows:
\textbf{(1)~Benchmark}:~A universal video retrieval benchmark with $16$ test datasets for the multi-dimensional, diagnostic capability evaluation.
\textbf{(2)~Data}:~A scalable video data synthesis workflow, producing over $1.55$ million cross-domain and multi-task pairs to establish a high-quality training resource.
\textbf{(3)~Training \& Model}:~A multimodal pyramid curriculum for learning generalizable video embeddings by modeling inherent knowledge dependencies across tasks.
\textbf{(4)~Experiment \& Analysis}:~Extensive experimental results and analysis, validating the superiority of our proposed methods among $14$ state-of-the-art video retrievers and discovering unnoticed and insightful knowledge.

\section{Related Works}

\paragraph{Video Retrieval.}
Text-to-video retrieval has progressed from matching coarse phrases to parsing fine-grained spatio-temporal descriptions~\cite{xu2016msr}. While recent benchmarks have advanced beyond simple recognition by incorporating detailed annotations like spatio-temporal grounding (CaReBench~\cite{xu2025carebench}), scene understanding (UltraVideo~\cite{ultravideo}), and camera motion (CameraBench~\cite{lin2025towards}), they remain specialized. Concurrently, the scope of retrieval has expanded to include new paradigms like composed queries with text-image pairs (e.g., CoVR~\cite{ventura2024covr,hummel2024egocvr}) and purely visual queries in egocentric contexts~\cite{liu2021activity}. Despite these advances, the field remains fragmented, with models and evaluation siloed within specific tasks or domains. This prevents a holistic understanding of a model's true generalization capabilities, a gap our unified benchmark aims to fill.

\paragraph{Video Embedding Models.}
Video embedding models have evolved from the adaptation of image-centric ones, such as CLIP~\cite{radford2021learning}, to powerful, larger language-based encoders. Early methods such as CLIP4Clip~\cite{luo2022clip4clip} and InternVideo2~\cite{wang2024internvideo2} added temporal modules to CLIP but inherited its limitations in complex language understanding and long-context processing~\cite{wang2025internvideo2,li2025improving}. To overcome these issues, recent work leverages Multimodal Large Language Models (MLLMs) as video embedders. Models like LLaVE~\cite{lan2025llave}, UNITE~\cite{kong2025modality}, and VLM2Vec-V2~\cite{meng2025vlm2vec} achieve strong performance on benchmarks by training on text-image and text-video data. However, this data has so far been narrow for learning generalizable embeddings, and these video embedding models have not been systematically evaluated across more complex tasks and domains of video retrieval.

\section{Methodology}

This section establishes a new ecosystem to reshape the fragmented scope of video retrieval by a co-designed, tripartite framework. The basis is the \textbf{Universal Video Retrieval Benchmark~(UVRB)}, which defines a comprehensive suite of abilities and serves as a diagnostic tool. Informed by UVRB's diagnostics, our \textbf{V-SynFlow} pipeline synthetically generates a high-fidelity dataset, \textbf{Universal Video Retrieval Dataset~(UVRD)}, engineered to populate the identified semantic and structural gaps. Finally, the \textbf{Modality Pyramid} provides a principled curriculum with adaptive task scheduling to train a \textbf{General Video Embedder~(GVE)}. This tight integration of diagnostic evaluation, targeted data synthesis, and model optimization forms a feasible solution for universal video retrieval.

\subsection{Unifying and Benchmarking Video Retrieval}\label{sec:uvrb}
Existing works are typically confined to coarse-grained text-to-video tasks, limiting their capacity to define and evaluate model generalization. To address this, we propose a new paradigm to unify complex query formats and divergent data domains of video retrieval, namely \textbf{Universal Video Retrieval~(UVR)}, defined as below.

\begin{definition}[Universal Video Retrieval~(UVR)]\label{def:universal_video_retrieval}
    Given a related pair with a query $q$ and a video $v$, UVR aims to learn a $\theta$-parameterized model $E_\theta(\cdot)$ to compute a relevance score between their embeddings, $s_{q \to v} = \cos(E_\theta(q), E_\theta(v))$, which should be higher than other irrelevant pairs. This condition can be satisfied for the given pair with different formats and in divergent domains.
    Specifically,
    \textbf{(1)~Query Format} can be Textual~(TXT, e.g., natural language), Composed~(CMP, text+image/text+video), Visual~(VIS, image/video).
    \textbf{(2)~Data Domain} can be Coarse-grained~(CG, high-level semantics), Fine-grained — Spatial~(S, object appearance), Temporal~(T, event dynamics), Partially Relevant~(PR, local or abstract information) — and Long-context~(LC, extended inputs).
\end{definition}

To systematically evaluate UVR performance, we introduce the \textbf{Universal Video Retrieval Benchmark (UVRB)}, which evaluates model universality by covering 16 test datasets targeting distinct abilities\footnote{We define an \textbf{ability} as proficiency in one query format \textit{or} one data domain (e.g., VIS is an ability; T is an ability). A general embedding model should master multiple abilities and their combinations.}.
Dataset statistics and construction details are presented in Appendix~\ref{app:uvrb}.
Specifically, coarse-grained tasks use MSRVTT~\cite{xu2016msr}, DiDeMo~\cite{anne2017localizing}, CRB-G~\cite{xu2025carebench}. Fine-grained: CRB-S~\cite{xu2025carebench}/VDC-O~\cite{chai2024auroracap} (spatial), CRB-T~\cite{xu2025carebench}/CMRB~\cite{lin2025towards} (temporal), DREAM-E~\cite{wang2024tarsier}/LoVR-Theme2Clip~\cite{cai2025lovr}/PEV-K~\cite{bolya2025perception} (partially relevant). Long-context: LoVR-V~\cite{cai2025lovr}, VDC-D~\cite{chai2024auroracap}. Composed queries: MS-TI/MS-TV (adapted from MomentSeeker~\cite{yuan2025momentseeker}), MSRVTT-I2V~\cite{xu2016msr}, LoVR-C2V~\cite{cai2025lovr}. 

To our knowledge, UVRB is the first benchmark to systematically span comprehensive video retrieval scenarios.
Through extensive experiments on UVRB across $14$ baselines in Section~\ref{sec:exp} to diagnose the strengths and weaknesses of existing works, which insightfully guide our design of data synthesis~(Section~\ref{sec:data_synthesis}) and model training~(Section~\ref{sec:modality_pyramid}).

\begin{figure}
    \centering
    \includegraphics[width=\linewidth]{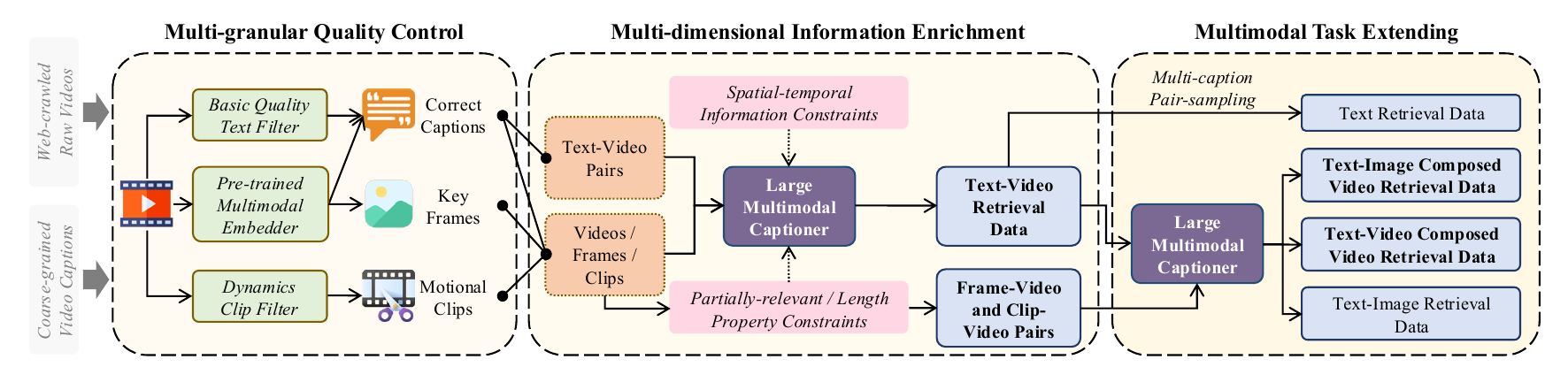}
    \caption{V-SynFlow: a multi-stage synthesis workflow for diverse video retrieval data.}
    \label{fig:v-synflow}
\end{figure}

\subsection{Scalable Synthesis of Cross-Domain Multi-Task Video Retrieval Data}\label{sec:data_synthesis}
Training a universal video embedder is fundamentally impeded by a deficiency of high-quality supervision for divergent tasks. Therefore, we introduce \textbf{V-SynFlow}~(Figure~\ref{fig:v-synflow}) to transform weakly annotated web videos from raw datasets~(e.g., PVD~\cite{bolya2025perception}, InternVid-FLT~\cite{wang2023internvid}, and WebVid~\cite{bain2021frozen}) into a structured, high-fidelity, multi-task training instances. In this way, we obtain a practical and diverse dataset, called \textbf{Universal Video Retrieval Dataset~(UVRD)} with over $1.55$ million descriptive pairs in total for dimensional training enhancement~(see Appendix~\ref{app:training_data} for details).
V-SynFlow proceeds in three stages:
We first construct a clean, semantically coherent material pool by filtering noise at multiple granularities.
Then we leverage an MLLM as a conditional generative engine to enrich semantic dimensions.
Lastly, we synthesize diverse instances across multiple retrieval tasks.
We provide our details and applied prompts for synthesis in Appendix~\ref{app:prompt}.

\paragraph{Multi-granular Quality Control.}
Given a raw corpus $\mathcal{D} = \{(v_i, t_i)\}$, we produce a high-fidelity asset pool, $\mathcal{A}_{tfc}$. The process applies a filter cascade: \textit{Annotation Rectification} to remove non-descriptive text; \textit{Cross-Modal Consistency Filtering}, which discards pairs where the similarity from a pretrained embedder $\Phi(\cdot)$~(e.g., GME-7B~\cite{Zhang_2025_CVPR}) is below a threshold; and \textit{Temporal Dynamics Filtering} to remove static content. The resulting asset pool $\mathcal{A}_{tfc}$ contains a set of validated videos $\{v_j\}$, their original captions $\{t_j\}$, and corresponding sets of extracted frames $\{f_{jk}\}$ and cropped clips $\{c_{jl}\}$.

\paragraph{Multi-dimensional Information Enrichment.}
We leverage the filtered assets in $\mathcal{A}_{tfc}$ to generate richer data structures.
To create an enriched text-video dataset $\mathcal{D}^{+}$, we use an MLLM, $\mathcal{M}_{\text{cap}}$~(e.g., Keye-VL-8B~\cite{team2025kwai}), as a conditional captioning engine.
For each video $v_j$, it synthesizes multiple captions $\{t^{'}_{jk}\}$ conditioned by randomly generated information profiles~($30\%$ spatial, $60\%$ temporal, and $10\%$ others).
By sampling one of the captions for each video, we obtain a set of new high-quality text-video pairs $\{(v_j, t^{'}_{jk})\}$.
Besides, we form pairs of a video and its visual components to construct collections of visual pairs $\mathcal{P}_{f \leftrightarrow v}$ and $\mathcal{P}_{c \leftrightarrow v}$, resulting in frame-to-video $\{(f_{jk}, v_j)\}$ and clip-to-video $\{(c_{jl}, v_j)\}$ pairs, respectively.

\paragraph{Multimodal Task Extension.}
The final stage assembles the unified training corpus, $\mathcal{D}_{\star}$.
We synthesize complex composed retrieval tasks by leveraging $\mathcal{P}_{f \leftrightarrow v}$ and $\mathcal{P}_{c \leftrightarrow v}$. For each visual pair (e.g., $(f_{jk}, v_j)$, $(c_{jl}, v_j)$), $\mathcal{M}_{\text{cap}}$ generates a query text $t^{f_{jk} \to v_j}$ describing the temporal evolution, forming a training instance $((t^{f_{jk} \to v_j}, f_{jk}), v_j)$.
Besides, basic alignment tasks (e.g., text-image pairs) are also sampled from $\mathcal{A}_{tfc}$. For each video, two candidates from the unselected synthesized captions are mapped as text-to-text pairs.
The resulting dataset $\mathcal{D}_{\star}$ provides a comprehensive mix of tasks essential for training a universal embedder.

\begin{figure}[t]
    \centering
    \begin{minipage}[b]{0.71\textwidth}
        \centering
        \includegraphics[width=\linewidth]{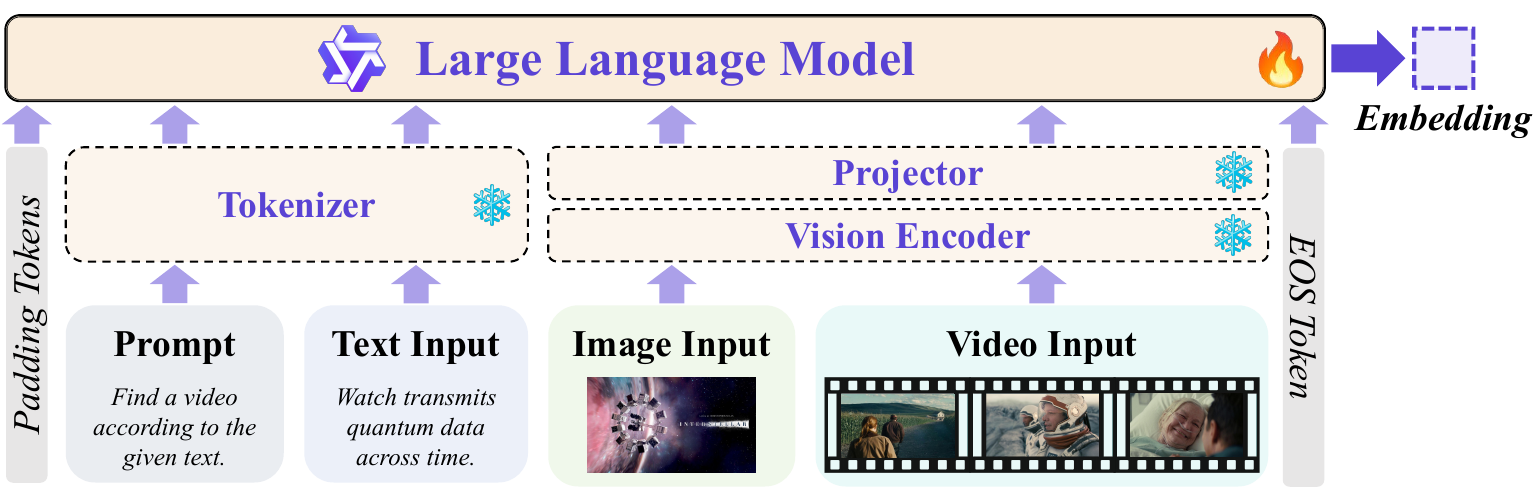}
        \captionof{figure}{The architecture of GVE, a MLLM-based embedding model. We only fine-tune the LLM part. GVE inputs compositional multimodal elements and outputs a high-dimensional vector as an embedding.}
        \label{fig:model}
    \end{minipage}
    \hfill
    \begin{minipage}[b]{0.28\textwidth}
        \centering
        \includegraphics[width=\linewidth]{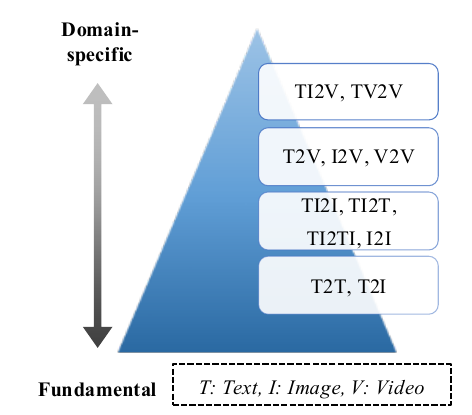}
        \captionof{figure}{Modality Pyramid: simpler tasks lay the foundation for specific ones.}
        \label{fig:pyramid}
    \end{minipage}
\end{figure}

\subsection{Modality Pyramid: Customized Curriculum Contrastive Learning for Generalizable Embeddings}
\label{sec:modality_pyramid}

Our synthesized dataset provides a rich mixture of retrieval tasks spanning diverse query formats (text, image, video, and their compositions) and data domains (coarse- to fine-grained, long-context, etc., see Appendix~\ref{app:training_data} for details). To embed multimodal inputs into a unified space, we introduce \textbf{General Video Embedder~(GVE)}~(Figure~\ref{fig:model}), a multimodal encoder derived from Qwen2.5-VL~\cite{bai2025qwen25vl} to inherit its pretrained vision-language aligned knowledge.
For arbitrary modality combinations (e.g., image-only or text+video), GVE fuses the tokenized prompt and text inputs with projected visual features into a joint input sequence. The LLM processes this sequence to autoregressively produce representations. We extract the final embedding via last token pooling and $\ell_2$-normalization for retrieval. Details are provided in Appendix~\ref{app:arch}.

However, our diagnostics based on UVRB reveal that naively training a single embedder on this heterogeneous data leads to suboptimal performance~(Section~\ref{sec:ablation}).
One of the potential reasons is that easy tasks dominate early optimization, while challenging ones receive insufficient gradient signal and converge poorly.
Moreover, existing methods either train on single-task data or overlook the knowledge dependencies between heterogeneous domains of multi-task data, which can benefit the joint incorporation of model abilities.
To address this, we propose \textsc{Modality Pyramid} (Figure~\ref{fig:pyramid}), a curriculum that schedules training from atomic to composite tasks for progressive knowledge acquisition. It guides the embedding model to master perceptual primitives first, then advance to complex integration, which preserves foundational knowledge while cultivating generalizable transfer.

\paragraph{Alignment-aware dynamic scheduling.}
At the beginning of each training epoch $t$, we estimate the alignment level of every task $k \in \mathcal{K}$ using a \textit{prober model} $\Psi_t$. For $t=1$, $\Psi_1$ is a strong off-the-shelf embedder (e.g., GME-7B~\cite{Zhang_2025_CVPR}); for $t>1$, $\Psi_t$ is the GVE checkpoint from the end of epoch $t-1$. For each task, we sample $N_p$ positive pairs and compute its relevance score as the average cosine similarity: $R_k^{(t)} = \frac{1}{N_p} \sum_i \cos(\Psi_t(x_i), \Psi_t(y_i))$. Higher $R_k^{(t)}$ indicates better current alignment.
During epoch $t$, tasks are sampled with probability $P^{(t)}(k) \propto \exp(R_k^{(t)} / \sigma(t))$, where the temperature $\sigma(t)$ increases linearly from $\sigma_{\min}$ to $\sigma_{\max}$~($\sigma_{\min}=0.1$ and $\sigma_{\max}=1.0$ by default). This annealing schedule ensures initial focus on well-aligned tasks while progressively encouraging exploration of challenging ones.

\paragraph{Unified Contrastive Optimization.}
GVE is trained with a symmetric InfoNCE loss~\cite{oord2018representation} across all scheduled tasks. To strengthen discrimination, we augment in-batch negatives with hard negatives mined from a large external corpus using the same prober $\Psi_t$. For a query–target pair $(q, v^+)$, the loss contrasts $v^+$ against both other positives in the batch and the top-$K$ retrieved hard negatives. The similarity is computed as $s_{q \to v} = \cos(E_\theta(q), E_\theta(v))$, and the final loss is symmetrized over query and target directions:
\begin{equation}
    \mathcal{L}_{i}^{(q \to y)} = -\log \frac{\exp(s_{q_i \to y_i^+} / \tau_l)}{\exp(s_{q_i \to y_i^+} / \tau_l) + \sum_{j \neq i} \exp(s_{q_i \to y_j^+} / \tau_l) + \sum_{y_k^- \in \mathcal{H}_i} \exp(s_{q_i \to y_k^-} / \tau_l)},
    \label{eq:infonce}
\end{equation}
where $s_{q \to v} = \cos(E_\theta(q), E_\theta(v))$ and $\tau_l$ is a pre-defined temperature. The total loss is the symmetric sum $\mathcal{L}_i = \frac{1}{2}(\mathcal{L}_i^{(q \to y)} + \mathcal{L}_i^{(y \to q)})$.

\section{Experiments}\label{sec:exp}

\subsection{Experimental Setups}\label{sec:setup}

\paragraph{Baselines.}
Our evaluation benchmarks $14$ prominent baselines. As detailed in Appendix~\ref{app:baseline_details}, our selection spans a wide range of architectures, parameter sizes (from $87$M to $8.3$B), and training data compositions. The models are broadly divided into two categories:
First, traditional \textit{CLIP-based} embedding models include \texttt{CLIP4Clip}~\cite{luo2022clip4clip}, \texttt{ViCLIP}~\cite{wang2023internvid}, \texttt{VideoCLIP-XL}~\cite{wang2024videoclip}, \texttt{LanguageBind}~\cite{zhu2024languagebind}, and the \texttt{InternVideo2} series ($1$B and $6$B)~\cite{wang2024internvideo2}.
Second, a more recent category of \textit{MLLM-based} embedding models includes \texttt{GME-2B}~\cite{Zhang_2025_CVPR}, \texttt{Unite-2B}~\cite{kong2025modality}, \texttt{VLM2Vec-V2}~\cite{meng2025vlm2vec}, \texttt{BGE-VL}~\cite{zhou2024megapairs}, \texttt{UniME-7B}~\cite{gu2025breaking}, \texttt{B3-7B}~\cite{thirukovalluru2025breaking}, \texttt{GME-7B}~\cite{Zhang_2025_CVPR}, and \texttt{Unite-7B}~\cite{kong2025modality}.
Note that the training data of baseline models may include in-domain data of test datasets in UVRB~(e.g, MSRVTT, DiDeMo).

\paragraph{Metrics.}
Our primary metric is Recall@1~(R@1), which measures if the most relevant item is the correct one. For more challenging datasets with fuzzy queries~(e.g., CMRB and LoVR-TH), we choose to report Recall@10~(R@10). Additionally, we use Precision@1~(P@1) for the MS-TI and MS-TV with multiple positive candidates.

\paragraph{Evaluation Implementations.}
All models are evaluated within UVRB under a controlled and consistent environment to ensure fairness and reproducibility. Model parameters are loaded in \texttt{bf16} precision wherever supported. All output embeddings are normalized to mitigate precision-induced variance. For retrieval tasks, we uniformly adopt cosine similarity as the relevance metric\footnote{Any learnable relevance estimation modules (e.g., the MLP head in InternVideo2) are removed and the last hidden state before them is used as embedding for fair comparison.}, with no post-processing or re-ranking applied. Only raw visual frames are used, and audio, speech, and metadata are excluded. Each video is uniformly sampled into exactly $8$ frames. CLIP-based models operate at $224 \times 224$ resolution, while MLLM-based models are constrained such that each frame encodes to fewer than $200$ visual tokens\footnote{Frame resolution is adaptively adjusted per model’s tokenization strategy to enforce the token limit.}. Input sequences for queries or candidates are capped at $8192$ tokens, with truncation applied beyond this limit. For models lacking native video support (e.g., BGE-VL), we implement a multi-image embedding pipeline~(usually by inserting multiple special tokens for image), treating each frame as an independent image input.

\paragraph{Others.} We provide complete details in the appendix, covering: (1) the construction and evaluation of UVRB~(Appendix~\ref{app:uvrb}-\ref{app:evaluation_pipeline}); (2) the data synthesis pipeline with prompts~(Appendix~\ref{app:training_data}-\ref{app:prompt}); and (3) the GVE model's architecture and training specifics~(Appendix~\ref{app:arch}-\ref{app:hyperparameters}). The appendix also includes baseline properties and more experimental results~(Appendix~\ref{app:baseline_details}-\ref{app:video_classification}).

\begin{table}[t]
\centering
\footnotesize
\setlength{\tabcolsep}{2.5pt}
\caption{Video retrieval performance for datasets of UVRB. The AVG values are averaged over $16$ datasets. For each column: highest score is \textbf{bolded}, second-highest is \underline{underlined}. Metrics: R@1 (Recall@1), R@10 (Recall@10), P@1 (Precision@1).}
\label{tab:main_by_datasets}
\begin{tabular}{l|>{\columncolor{gray!15}}c|*{8}{c}}
\toprule
\textbf{Model} & \textbf{AVG} & \textbf{MSRVTT} & \textbf{DiDeMo} & \textbf{CRB-G} & \textbf{CRB-S} & \textbf{VDC-O} & \textbf{CRB-T} & \textbf{CMRB} & \textbf{DREAM-E} \\
& & R@1 & R@1 & R@1 & R@1 & R@1 & R@1 & R@10 & R@1 \\
\midrule
CLIP4Clip & 0.390 & 0.333 & 0.297 & 0.511 & 0.497 & 0.620 & 0.289 & 0.280 & 0.191 \\
ViCLIP & 0.352 & 0.386 & 0.306 & 0.447 & 0.437 & 0.530 & 0.349 & 0.229 & 0.235 \\
VideoCLIP-XL & 0.491 & 0.443 & 0.403 & 0.828 & 0.839 & 0.735 & 0.487 & 0.274 & 0.263 \\
LanguageBind & 0.487 & \underline{0.479} & \underline{0.421} & 0.716 & 0.687 & 0.759 & 0.466 & 0.290 & 0.280 \\
InternVideo2-1B & 0.404 & 0.449 & 0.404 & 0.586 & 0.568 & 0.644 & 0.470 & 0.355 & 0.242 \\
InternVideo2-6B & 0.427 & \textbf{0.485} & 0.418 & 0.608 & 0.612 & 0.650 & 0.455 & 0.346 & 0.271 \\
GME-2B & 0.488 & 0.390 & 0.303 & 0.690 & 0.718 & 0.715 & 0.400 & 0.298 & 0.240 \\
Unite-2B & 0.480 & 0.367 & 0.298 & 0.699 & 0.723 & 0.727 & 0.409 & 0.284 & 0.223 \\
VLM2Vec-V2 & 0.508 & 0.330 & 0.299 & 0.828 & 0.843 & 0.775 & 0.410 & 0.286 & 0.228 \\
BGE-VL & 0.443 & 0.337 & 0.318 & 0.690 & 0.688 & 0.639 & 0.359 & 0.225 & 0.212 \\
UniME-7B & 0.521 & 0.351 & 0.335 & 0.815 & 0.827 & 0.743 & 0.476 & 0.317 & \underline{0.293} \\
B3-7B & 0.511 & 0.282 & 0.350 & 0.815 & 0.825 & 0.768 & 0.415 & 0.312 & 0.216 \\
GME-7B & 0.530 & 0.436 & 0.377 & 0.740 & 0.767 & 0.731 & 0.442 & 0.304 & 0.274 \\
Unite-7B & 0.538 & 0.439 & 0.386 & 0.798 & 0.804 & 0.753 & 0.472 & 0.351 & 0.279 \\
\midrule
GVE-3B & \underline{0.544} & 0.431 & 0.376 & \underline{0.850} & \underline{0.846} & \underline{0.786} & \underline{0.496} & \underline{0.363} & 0.280 \\
GVE-7B & \textbf{0.573} & 0.464 & \textbf{0.433} & \textbf{0.865} & \textbf{0.847} & \textbf{0.794} & \textbf{0.539} & \textbf{0.398} & \textbf{0.302} \\
\bottomrule
\end{tabular}

\vspace{5pt}

\begin{tabular}{l|*{8}{c}}
\toprule
\textbf{Model} & \textbf{LoVR-TH} & \textbf{PEV-K} & \textbf{LoVR-V} & \textbf{VDC-D} & \textbf{MS-TI} & \textbf{MS-TV} & \textbf{MSRVTT-I2V} & \textbf{LoVR-C2V} \\
& R@10 & R@1 & R@1 & R@1 & P@1 & P@1 & R@1 & R@1 \\
\midrule
CLIP4Clip & 0.338 & 0.179 & 0.360 & 0.566 & 0.173 & 0.183 & \textbf{0.924} & \underline{0.503} \\
ViCLIP & 0.202 & 0.075 & 0.230 & 0.395 & 0.283 & 0.243 & 0.846 & 0.433 \\
VideoCLIP-XL & 0.439 & 0.229 & 0.380 & 0.820 & 0.230 & 0.223 & 0.861 & 0.403 \\
LanguageBind & 0.425 & 0.303 & 0.540 & 0.679 & 0.228 & 0.233 & 0.827 & 0.463 \\
InternVideo2-1B & 0.298 & 0.026 & 0.280 & 0.485 & 0.265 & 0.230 & 0.794 & 0.368 \\
InternVideo2-6B & 0.302 & 0.086 & 0.330 & 0.516 & 0.235 & 0.205 & 0.868 & 0.452 \\
GME-2B & 0.446 & 0.354 & 0.530 & 0.839 & \textbf{0.350} & \textbf{0.340} & 0.827 & 0.366 \\
Unite-2B & 0.445 & 0.355 & 0.570 & 0.792 & 0.250 & 0.233 & 0.863 & 0.445 \\
VLM2Vec-V2 & 0.492 & 0.324 & 0.610 & 0.913 & 0.275 & 0.250 & 0.841 & 0.385 \\
BGE-VL & 0.387 & 0.184 & 0.550 & 0.722 & 0.303 & 0.233 & 0.779 & 0.465 \\
UniME-7B & 0.504 & 0.323 & 0.480 & 0.847 & 0.310 & 0.305 & 0.867 & \textbf{0.537} \\
B3-7B & 0.462 & 0.387 & 0.590 & 0.853 & 0.275 & 0.265 & 0.884 & 0.471 \\
GME-7B & 0.523 & 0.396 & \textbf{0.710} & 0.865 & \underline{0.348} & \underline{0.333} & 0.860 & 0.370 \\
Unite-7B & \textbf{0.555} & \textbf{0.440} & 0.620 & 0.871 & 0.278 & 0.230 & 0.883 & 0.448 \\
\midrule
GVE-3B & 0.522 & 0.330 & 0.610 & \underline{0.918} & 0.340 & 0.268 & 0.891 & 0.403 \\
GVE-7B & \underline{0.542} & \underline{0.413} & \underline{0.680} & \textbf{0.948} & 0.343 & 0.280 & \underline{0.899} & 0.415 \\
\bottomrule
\end{tabular}
\end{table}

\begin{table}[t]

\centering
\footnotesize
\caption{Video retrieval performance by specific abilities~(tasks and domains) on UVRB. The AVG values are averaged over tasks~(textual~(TXT), composed~(CMP), visual~(VIS)) and domains~(coarse-grained~(CG), fine-grained~(FG), long-context~(LC)) video retrieval tasks. Besides, we provide sub-domain results, including spatial~(S), temporal~(T), partially~relevant~(PR). For each column: highest score is \textbf{bolded}, second-highest is \underline{underlined}.}
\label{tab:main_by_abilities}
\begin{tabular}{l|>{\columncolor{gray!40}}c|ccc|ccc|ccc}
\toprule
& & \multicolumn{3}{c|}{\textbf{Tasks}} & \multicolumn{3}{c|}{\textbf{Domains}} & \multicolumn{3}{c}{\textbf{Sub-domains}} \\
\midrule
\textbf{Model} & \textbf{AVG} & \textbf{TXT} & \textbf{CMP} & \textbf{VIS} & \textbf{CG} & \textbf{FG} & \textbf{LC} & \textbf{S} & \textbf{T} & \textbf{PR} \\
\midrule
CLIP4Clip        & 0.416 & 0.401 & 0.178 & \textbf{0.714} & 0.380 & 0.360 & 0.463 & 0.559 & 0.285 & 0.236 \\
ViCLIP           & 0.375 & 0.336 & 0.263 & 0.640 & 0.380 & 0.315 & 0.313 & 0.484 & 0.289 & 0.171 \\
VideoCLIP-XL     & 0.510 & 0.550 & 0.227 & 0.632 & \underline{0.558} & 0.493 & 0.600 & 0.787 & 0.381 & 0.310 \\
LanguageBind     & 0.508 & 0.543 & 0.231 & 0.645 & 0.539 & 0.479 & 0.610 & 0.723 & 0.378 & 0.336 \\
InternVideo2-1B  & 0.420 & 0.422 & 0.248 & 0.581 & 0.480 & 0.403 & 0.383 & 0.606 & 0.413 & 0.189 \\
InternVideo2-6B  & 0.445 & 0.448 & 0.220 & 0.660 & 0.504 & 0.417 & 0.423 & 0.631 & 0.400 & 0.220 \\
GME-2B           & 0.416 & 0.539 & \textbf{0.345} & 0.597 & 0.461 & 0.471 & 0.685 & 0.716 & 0.349 & 0.347 \\
Unite-2B         & 0.507 & 0.536 & 0.242 & 0.654 & 0.455 & 0.471 & 0.681 & 0.725 & 0.347 & 0.341 \\
VLM2Vec-V2       & 0.538 & 0.587 & 0.263 & 0.613 & 0.498 & 0.502 & 0.762 & 0.809 & 0.348 & 0.348 \\
BGE-VL           & 0.480 & 0.497 & 0.268 & 0.622 & 0.448 & 0.406 & 0.636 & 0.664 & 0.292 & 0.261 \\
UniME-7B         & 0.542 & 0.561 & 0.308 & \underline{0.702} & 0.500 & 0.518 & 0.664 & 0.785 & 0.396 & 0.373 \\
B3-7B            & 0.538 & 0.570 & 0.270 & 0.678 & 0.482 & 0.505 & 0.722 & 0.797 & 0.364 & 0.355 \\
GME-7B           & 0.562 & 0.604 & \underline{0.341} & 0.615 & 0.518 & 0.507 & \underline{0.788} & 0.749 & 0.373 & 0.398 \\
Unite-7B         & 0.559 & 0.609 & 0.254 & 0.666 & 0.541 & 0.539 & 0.746 & 0.779 & 0.412 & \textbf{0.425} \\
\midrule
GVE-3B           & \underline{0.571} & \underline{0.619} & 0.304 & 0.647 & 0.552 & \underline{0.541} & 0.764 & \underline{0.816} & \underline{0.430} & 0.377 \\
GVE-7B           & \textbf{0.600} & \textbf{0.657} & 0.312 & 0.657 & \textbf{0.587} & \textbf{0.570} & \textbf{0.814} & \textbf{0.821} & \textbf{0.469} & \underline{0.419} \\
\bottomrule
\end{tabular}
\end{table}

\subsection{Main Results}\label{sec:main_results}

\paragraph{Overall Performance.}
We evaluate \texttt{GVE} on the UVRB benchmark under a strictly zero-shot setting without any exposure to in-domain data during training. Although competing models may have an unfair advantage for using training data corresponding to several test sets, our results in Table~\ref{tab:main_by_datasets} and Table~\ref{tab:main_by_abilities} show clear superiority and confirm the strong generalization of \texttt{GVE}.
Specifically, \texttt{GVE-7B} achieves state-of-the-art results with mean scores of $0.573$ across datasets and $0.600$ across task categories. It outperforms \texttt{Unite-7B} by $+6.5\%$ and $+7.3\%$ even though \texttt{Unite-7B} may have seen in-domain data. \texttt{GVE-7B} leads in every major dimension including TXT at $0.657$ versus $0.609$, CMP at $0.312$ versus $0.254$, CG at $0.587$, FG at $0.570$, and LC at $0.814$ versus $0.746$. It also leads in fine-grained subdomains with S at $0.821$ versus $0.779$ and T at $0.469$ versus $0.412$. \texttt{Unite-7B} shows strength in VIS and PR tasks but underperforms in compositional and temporal reasoning. Its performance is uneven and relies heavily on specific training data. The compact \texttt{GVE-3B} with $3.8$B parameters scores $0.571$, higher than \texttt{Unite-7B} at $0.559$ with over $7$B parameters. This shows our gains come from better data synthesis and curriculum design not from model size or data leakage. Our smaller models match or beat larger ones under fair zero-shot evaluation. This advantage translates to top results while \texttt{Unite-7B}'s strength on a few datasets reflects narrow capability.

\begin{table}[t]
\centering
\footnotesize
\caption{Ablation study for synthesized data and modality pyramid curriculum. \colorbox{gray!15}{\textbf{AVG of D}}: average across datasets, \colorbox{gray!40}{\textbf{AVG of A}}: average across abilities. For each column of each size of model~(3B or 7B): highest score is \textbf{bolded}, second-highest is \underline{underlined}.}
\label{tab:ablation}
\begin{adjustbox}{width=\linewidth,center}
\begin{tabular}{l|>{\columncolor{gray!15}}c|>{\columncolor{gray!40}}c|ccc|ccc|ccc}
\toprule
& \textbf{D} & \textbf{A} & \multicolumn{3}{c|}{\textbf{Tasks}} & \multicolumn{3}{c|}{\textbf{Domains}} & \multicolumn{3}{c}{\textbf{Sub-domains}} \\
\midrule
\textbf{Model} & \textbf{AVG} & \textbf{AVG} & \textbf{TXT} & \textbf{CMP} & \textbf{VIS} & \textbf{CG} & \textbf{FG} & \textbf{LC} & \textbf{S} & \textbf{T} & \textbf{PR} \\
\midrule

GVE-i-3B & 0.528 & 0.558 & \textbf{0.620} & 0.237 & \underline{0.632} & 0.532 & 0.521 & \textbf{0.808} & 0.781 & 0.402 & \textbf{0.379} \\
GVE-s-3B & \underline{0.537} & \underline{0.564} & 0.617 & \underline{0.301} & 0.617 & \underline{0.539} & \underline{0.536} & \underline{0.775} & \underline{0.811} & \underline{0.421} & \underline{0.377} \\
GVE-3B & \textbf{0.544} & \textbf{0.571} & \underline{0.619} & \textbf{0.304} & \textbf{0.647} & \textbf{0.552} & \textbf{0.541} & 0.764 & \textbf{0.816} & \textbf{0.430} & \underline{0.377} \\

\midrule

GVE-i-7B & 0.563 & 0.587 & 0.643 & 0.274 & \textbf{0.678} & 0.567 & \underline{0.566} & 0.795 & 0.812 & \underline{0.459} & \textbf{0.426} \\
GVE-s-7B & \underline{0.568} & \underline{0.594} & \underline{0.648} & \textbf{0.313} & \underline{0.662} & \underline{0.576} & 0.563 & \underline{0.804} & \underline{0.814} & 0.458 & 0.418 \\
GVE-7B & \textbf{0.573} & \textbf{0.600} & \textbf{0.657} & \underline{0.312} & 0.657 & \textbf{0.587} & \textbf{0.570} & \textbf{0.814} & \textbf{0.821} & \textbf{0.469} & \underline{0.419} \\

\bottomrule
\end{tabular}
\end{adjustbox}
\end{table}

\paragraph{Ablation Study.}\label{sec:ablation}
We perform an ablation study to isolate the distinct contributions of our synthesized dataset (UVRD) and the Modality Pyramid curriculum, with results presented in Table~\ref{tab:ablation}. First, integrating UVRD (\texttt{GVE-s} vs. baseline \texttt{GVE-i}) demonstrates a significant impact on acquiring complex abilities; notably, performance on the composed (CMP) task for the 3B model increases by $27\%$ relative. Second, applying the Modality Pyramid (\texttt{GVE} vs. \texttt{GVE-s}) further refines model capabilities and enhances generalization, boosting the overall score from $0.594$ to $0.600$ and TXT score from $0.648$ to $0.657$ for the 7B model. In conclusion, UVRD provides the diverse knowledge required for complex tasks, while our curriculum optimizes its integration, culminating in a total performance gain of $1.8\%$ to $3.1\%$ over the baseline.

\begin{figure}[t]
    \centering
    \includegraphics[width=0.6\linewidth]{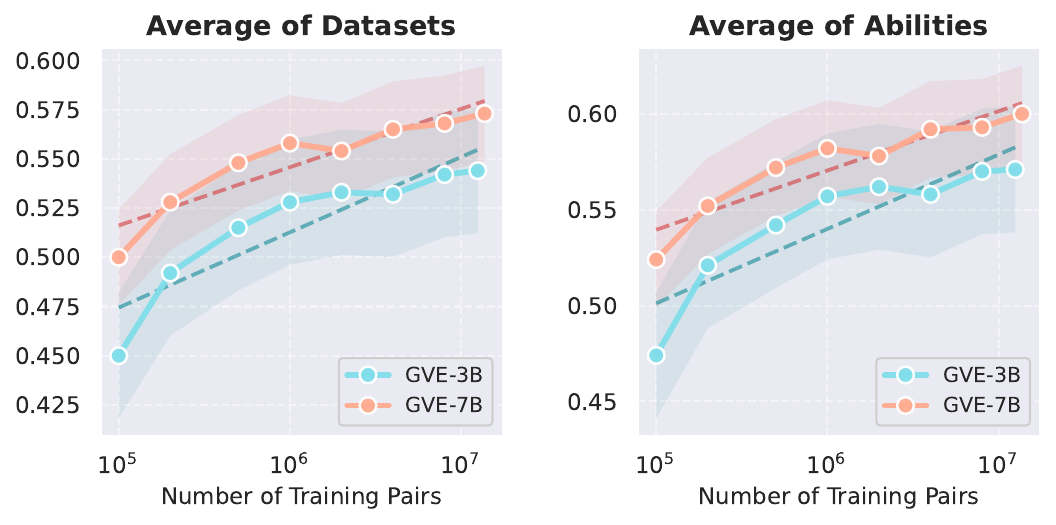}
    \caption{Performance effect from data scaling for GVE series. See Appendix~\ref{app:data_scaling} for detailed results.}
    \label{fig:scaling_avg}
\end{figure}

\paragraph{Data Scaling.}\label{sec:data_scaling}
Figure~\ref{fig:scaling_avg} shows the training-time scaling behavior that performance logarithmically improves with more data, but with diminishing returns.
We quantify scaling efficiency by fitting a logarithmic model $y = a \ln x + b$~($x$: data size, $y$: performance) and report the absolute and relative gain per $10\times$ data increase. 
On average across datasets~(abilities), \texttt{GVE-3B} improves by $+7.4\%$~($+7.1\%$) per decade, while \texttt{GVE-7B} gains $+5.4\%$~($+5.4\%$).
While \texttt{GVE-3B} exhibits higher scaling efficiency in relative terms, \texttt{GVE-7B} starts from a higher baseline, suggesting a trade-off between scaling slope and absolute capability.
This result highlights the potential for scaling a more powerful video embedding model with training data at a larger scale.
In addition, we also explore the test-time scaling of spatial and temporal density~(the frame number and resolution) in Appendix~\ref{app:scaling} as a dimension of generalization.

\subsection{Analysis of Dimensional Capabilities}
\label{sec:analysis}

\begin{figure}[t]
    \centering
    \includegraphics[width=\linewidth]{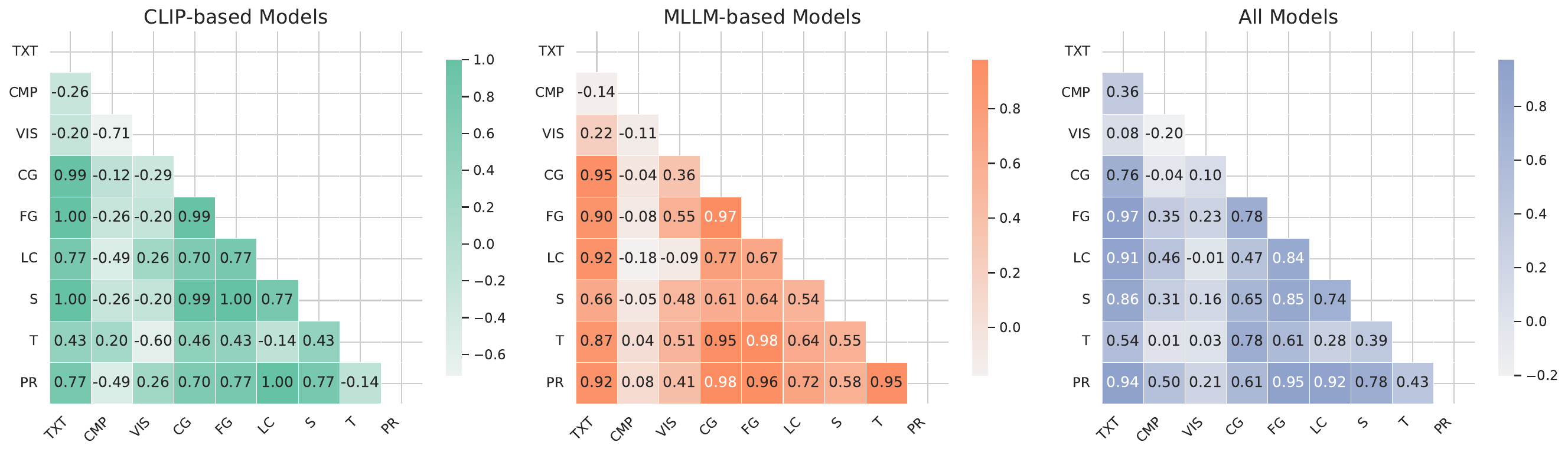}
    \caption{Correlation between dimensional abilities on UVRB for CLIP or MLLM-based models.}
    \label{fig:corr_ability}
\end{figure}

\begin{figure}[t]
    \centering
    \includegraphics[width=\linewidth]{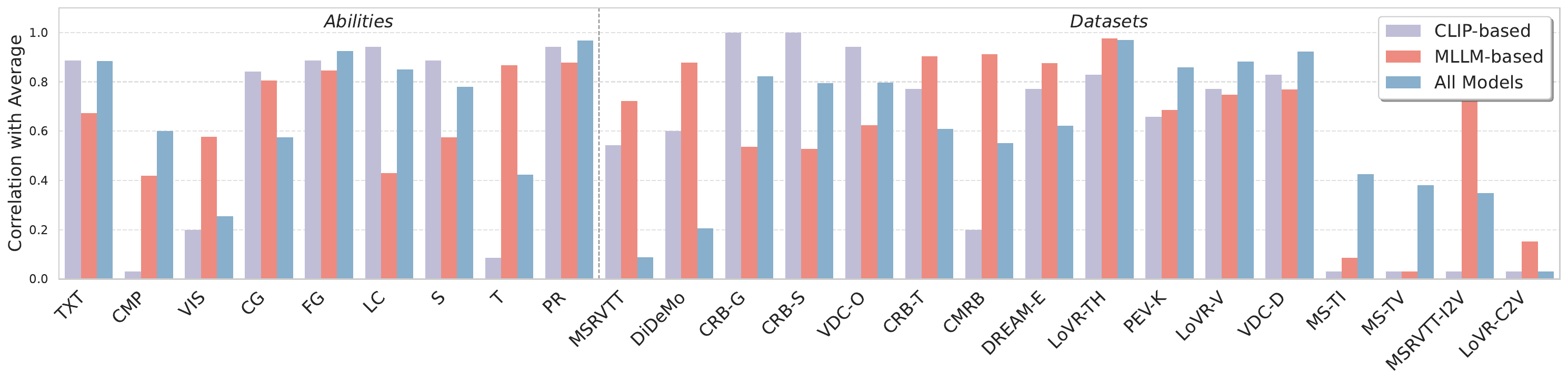}
    \caption{Correlation between averaged performance and abilities or datasets on UVRB.}
    \label{fig:corr_task_data}
\end{figure}

Figures~\ref{fig:corr_ability} and~\ref{fig:corr_task_data} reveal patterns in how video retrieval models develop capabilities and what their relationships are, as measured primarily by Pearson correlation ($\rho$) for their performance results. Here, we present several discoveries from the analysis in underexplored perspectives.

\paragraph{Finding 1: Partially Relevant Video Retrieval Better Reflects Universality than Traditional Benchmarks.}
Standard benchmarks such as MSRVTT show a low correlation with average UVR performance ($\rho_{\text{avg}} = 0.58$), suggesting limited representativeness, likely due to overfitting or simplified task design. In contrast, fine-grained (FG), partially relevant (PR), and long-context (LC) retrieval exhibit strong mutual correlations ($\rho \geq 0.90$). More importantly, PR retrieval, although understudied, achieves the highest average correlation with overall performance ($\rho_{\text{avg}} = 0.97$), positioning it as a promising proxy for robust model evaluation.

\paragraph{Finding 2: Disentangled Spatial and Temporal Representations.}
Models exhibit a marked decoupling between spatial (S) and temporal (T) representation ($\rho = 0.12$). This asymmetry is critical that temporal skills dominate fine-grained understanding ($\rho_{\text{T-FG}} = 0.98$), while spatial skills contribute minimally ($\rho_{\text{S-FG}} = 0.39$). This suggests current works fail to jointly model \textit{when} and \textit{where}, highlighting the need for inductive biases that encourage spatiotemporal integration.

\paragraph{Finding 3: Performance Divergence Between CLIP and MLLM-based Models.}
Model failures are architecture-dependent. For example, CLIP-based models are spatially biased ($\rho_{\text{S-CG}} = 0.99$) but temporally weak ($\rho_{\text{T-CG}} = 0.46$), leading to ability trade-offs. Compositional representation inversely correlates with visual accuracy ($\rho_{\text{CMP-VIS}} = -0.71$). And a near-zero link between temporal and long-context skills ($\rho_{\text{T-LC}} = -0.14$). In contrast, MLLM-based models demonstrate more balanced and integrated learning: superior semantic matching (PR-CG: MLLM $\rho = 0.98$ vs. CLIP $0.70$) and stronger temporal-long-context coupling ($\rho_{\text{LC-T}} = 0.64$). Therefore, architecture fundamentally shapes the development of capability, and MLLM has been increasingly popular because of its generalizability in video embedding modeling.

\paragraph{Finding 4: Scaling Has Limited Impact on Visual Perception.}
Parameter scaling improves high-level semantic coherence, but yields negligible gains in low-level visual perception. Notably, the $87$M-parameter \texttt{CLIP4Clip} (VIS: $0.714$) outperforms the $8$B-parameter \texttt{Unite-7B} (VIS: $0.702$). Given the weak correlation between visual fidelity and overall retrieval success ($\rho_{\text{AVG-VIS}} = 0.26$), future progress requires targeted improvements in visual grounding.

\section{Conclusions}
This work pioneers a unified paradigm for video retrieval. We introduced the first benchmark to comprehensively evaluate dimensional video retrieval abilities. It provides diagnostics to guide us to generate 1.55 million high-fidelity, multi-task training pairs to meet real-world complexity. In addition, we propose a novel curriculum learning algorithm to take advantage of the inherent task-wise relational structure. Based on these, we train a superior MLLM-based video embedding model, GVE.
Our experiments validate the state-of-the-art generalization of GVE on UVRB and provide insightful findings in this field.
Overall, this paper provides a foundational framework with evaluation-data-training co-design toward a more robust, versatile, and generalizable video retriever.



\bibliography{references}
\bibliographystyle{assets/plainnat}

\appendix
\section{Appendix}

\startcontents
\printcontents{}{1}{\subsection*{Content}}

\subsection{Overview of Appendix}\label{app:overview}
This appendix provides supplementary material supporting the main paper, organized as follows:

\paragraph{Methodology Supplementary.}
\begin{itemize}
    \item \textbf{Details of Benchmarking~(for Section~\ref{sec:uvrb})}: (1)~UVRB Details~(Appendix~\ref{app:uvrb}): Provides statistics, construction strategies, and evaluation details for the UVRB datasets. This includes definitions for coarse-grained, fine-grained (spatial, temporal, partially relevant), long-context, composed, and visual video retrieval tasks.
    (2)~Evaluation Pipeline~(Appendix~\ref{app:evaluation_pipeline}): Offers an overview of the scalable, reproducible evaluation protocol.
    \item \textbf{Details of Data Synthesis and Preparation~(for Section~\ref{sec:data_synthesis})}:
    (1)~Training Data~(Appendix~\ref{app:training_data}): Details the composition and scaling of the multi-modal training mixture, including descriptions of specific datasets used.
    (2)~Dataset Construction Pipeline~(Appendix~\ref{app:data_construct}): Explains the design principles (modularity, integrity) of the unified retrieval dataset framework.
    (3)~Synthetic Data Prompts~(Appendix~\ref{app:prompt}): Presents the structured prompts used for generating synthetic video retrieval data (covering captioning, composed retrieval, and frame-level tasks).
    \item \textbf{Details of Model and Training~(for Section~\ref{sec:modality_pyramid})}:
    (1)~Model Architecture~(Appendix~\ref{app:arch}): Gives a detailed description of the GVE model derived from Qwen2.5-VL, including input fusion and embedding extraction.
    (2)~Training Implementation~(Appendix~\ref{app:hyperparameters}): Outlines the parameter-efficient fine-tuning (LoRA) strategy, key hyperparameters, and the contrastive learning setup.
\end{itemize}

\paragraph{Experimental Details and Results.}
\begin{itemize}
    \item \textbf{Baseline Details (Appendix~\ref{app:baseline_details}):} Lists specifications of the evaluated baseline models.
    \item \textbf{Training Dynamics (Appendix~\ref{app:training_dynamics}):} Shows dynamics, status via metrics in training for GVE-3B and GVE-7B models.
    \item \textbf{More Experiments}:
    (1)~Training-time Scaling~(Appendix~\ref{app:data_scaling}): Presents the impact of scaling training data on performance in addition to Section~\ref{sec:data_scaling}.
    (2)~Test-time Scaling~(Appendix~\ref{app:scaling}): Analyzes the effect of scaling test-time parameters (number of frames, resolution) on performance extended from Section~\ref{sec:data_scaling}.
    (3)~Video Classification~(Appendix~\ref{app:video_classification}): Reports model performance on standard video classification benchmarks.
\end{itemize}

\paragraph{Others.}
\begin{itemize}
    \item \textbf{Limitations~(Appendix~\ref{app:limitations})}: Claim existing limitations about scopes and settings, required to be addressed or extended in future works.
\end{itemize}

\subsection{UVRB Details}\label{app:uvrb}

This section provides details of UVRB, including statistics and construction strategies of datasets.

First, the statistics for the number of queries and corpus, the video durations, and the text lengths are presented in Table~\ref{tab:uvrb_dataset}.

\begin{table}[ht!]
\centering
\footnotesize
\caption{Statistics of datasets in the Universal Video Retrieval Benchmark (UVRB). All videos use 8 uniformly sampled frames. $\#$~Query: the number of queries; $\#$~Corpus: the number of corpus; Dur~(s):~Duration in seconds; $\#$~Word:~text length in words.}
\label{tab:uvrb_dataset}
\begin{tabular}{l r r r r}
\toprule
\textbf{Dataset} & \textbf{$\#$~Query} & \textbf{$\#$~Corpus} & \textbf{Dur~(s)} & \textbf{$\#$~Word} \\
\midrule
\multicolumn{5}{l}{\textbf{Textual Video Retrieval~(Coarse-grained)}} \\
MSRVTT~\cite{xu2016msr} & 1,000 & 1,000 & 15.0 & 9.4 \\
DiDeMo~\cite{anne2017localizing} & 1,004 & 1,004 & 53.9 & 29.1 \\
CaReBench-General (CRB-G)~\cite{xu2025carebench} & 1,000 & 1,000 & 14.4 & 232.2 \\
\midrule
\multicolumn{5}{l}{\textbf{Textual Video Retrieval~(Fine-grained)}} \\
\multicolumn{5}{l}{\textbf{\textit{(a)~Spatial}}} \\
CaReBench-Spatial (CRB-S)~\cite{xu2025carebench} & 1,000 & 1,000 & 14.4 & 115.0 \\
VDC-Object~(VDC-O)~\cite{chai2024auroracap} & 1,027 & 1,027 & 30.1 & 91.4 \\
\multicolumn{5}{l}{\textbf{\textit{(b)~Temporal}}} \\
CaReBench-Temporal~(CRB-T)~\cite{xu2025carebench} & 1,000 & 1,000 & 14.4 & 103.2 \\
CameraBench~(CMRB)~\cite{lin2025towards} & 728 & 1,071 & 5.7 & 24.8 \\
\multicolumn{5}{l}{\textbf{\textit{(c)~Partially~Relevant}}} \\
DREAM-1K-Event~(DREAM-E)~\cite{wang2024tarsier} & 6,251 & 1,000 & 8.8 & 6.5 \\
LoVR-Theme2Clip~(LoVR-TH)~\cite{cai2025lovr} & 8,854 & 8,854 & 16.9 & 48.1 \\
PE-Video-Keyword~(PEV-K)~\cite{bolya2025perception} & 14,427 & 15,000 & 16.9 & 45.5 \\
\midrule
\multicolumn{5}{l}{\textbf{Textual Video Retrieval~(Long-context)}} \\
LoVR-Text2Video~(LoVR-V)~\cite{cai2025lovr} & 100 & 467 & 1560.3 & 17364.5 \\
VDC-Detail~(VDC-D)~\cite{chai2024auroracap} & 1,000 & 1,027 & 30.1 & 508.0 \\
\midrule
\multicolumn{5}{l}{\textbf{Composed Video Retrieval}} \\
MomentSeeker-Text-Image~(MS-TI)~\cite{yuan2025momentseeker} & 400 & 10 & 13.5 & 68.5 \\
MomentSeeker-Text-Video~(MS-TV)~\cite{yuan2025momentseeker} & 400 & 10 & 13.5 & 68.5 \\
\midrule
\multicolumn{5}{l}{\textbf{Visual Video Retrieval}} \\
MSRVTT-ImageVideo (MSRVTT-I2V)~\cite{xu2016msr} & 1,000 & 1,000 & 15.0 & - \\
LoVR-Clip-to-Video (LoVR-C2V)~\cite{cai2025lovr} & 467 & 467 & 1560.3 & - \\
\bottomrule
\end{tabular}
\end{table}

Second, we introduce the construction strategies of each dataset as follows.
\begin{itemize}
    \item~\textbf{MSRVTT:} We follow the data partition from JSFusion~\cite{yu2018joint}, utilizing 1,000 clip-text pairs from the original MSRVTT dataset~\cite{xu2016msr} for evaluation.

    \item~\textbf{DiDeMo:} Following the methodology in~\cite{liu2019use}, we concatenate all sentence descriptions associated with a single video from the DiDeMo dataset~\cite{anne2017localizing} to create a paragraph-level query for paragraph-to-video retrieval.

    \item~\textbf{CRB-G:} We adhere to the CaReBench protocol~\cite{xu2025carebench} and use the content from the \texttt{caption} field as the general query to retrieve videos.

    \item~\textbf{CRB-S:} Similar to CRB-G, we follow~\cite{xu2025carebench} and select the text from the \texttt{spatial\_caption} field to form queries focused on spatial descriptions.

    \item~\textbf{VDC-O:} We utilize the VDC dataset~\cite{chai2024auroracap} and extract text from the \texttt{main\_object\_caption} field as an object-centric query (e.g., \textit{The main subject, a worker dressed in a gray sleeveless shirt and beige pants...}).

    \item~\textbf{CRB-T:} Similar to CRB-S, we again follow~\cite{xu2025carebench}, using the \texttt{temporal\_caption} field to create queries based on temporal progression.

    \item~\textbf{CMRB:} We use the detailed camera motion annotations from CameraBench~\cite{lin2025towards} as queries to retrieve videos (e.g., \textit{The camera smoothly dollies forward, maintaining a steady and fluid motion...}).

    \item~\textbf{DREAM-E:} For event-based retrieval, we collect event descriptions from the DREAM-1K dataset~\cite{wang2024tarsier} to serve as queries for event-to-video matching (e.g., \textit{Wooden trap launches purple squirrel into the air}).

    \item~\textbf{LoVR-TH:} From the LoVR dataset~\cite{cai2025lovr}, we select the theme annotations of video clips as queries for theme-to-clip retrieval (e.g., \textit{The overall style of the animation is vibrant and whimsical...}).

    \item~\textbf{PEV-K:} We use the annotations from the \texttt{keyword} field in the PE-Video test set~\cite{bolya2025perception} to perform keyword-based partially relevant matching (e.g., \textit{colorful, paper, beautiful...}).

    \item~\textbf{LoVR-V:} We leverage the full-length video captions from the LoVR dataset~\cite{cai2025lovr} to perform long-text to long-video retrieval.

    \item~\textbf{VDC-D:} For this task, we extract the long, detailed descriptions from the \texttt{detailed\_caption} field of the VDC dataset~\cite{chai2024auroracap} to serve as fine-grained queries.

    \item~\textbf{MS-TI:} Following the adaptation method in~\cite{meng2025vlm2vec} for the MomentSeeker dataset~\cite{yuan2025momentseeker}, we create a text-image composed retrieval task. The goal is to retrieve a target video clip using a combined text and image query.

    \item~\textbf{MS-TV:} Similar to MS-TI, the task for MS-TV is to use a composed query of a text description and a reference video clip to retrieve the target clip.

    \item~\textbf{MSRVTT-I2V:} We construct image-to-video retrieval pairs from the aforementioned MSRVTT test set. For each video, a single frame is randomly sampled to serve as the image query for retrieving its source video.

    \item~\textbf{LoVR-C2V:} We leverage the original clip-to-video structure of the LoVR dataset~\cite{cai2025lovr}. For each full-length video, one of its corresponding short clips is used as the query for clip-to-video retrieval.
\end{itemize}

Third, we list the query prompt and metrics for datasets in evaluations in Table~\ref{tab:uvrb_prompts_metric}.

\begin{table}[ht!]
\centering
\footnotesize
\caption{Query Prompts and Metrics for Datasets in UVRB.}
\label{tab:uvrb_prompts_metric}
\begin{tabular}{l p{9cm} l}
\toprule
\textbf{Dataset} & \textbf{Query Prompt} & \textbf{Metric} \\
\midrule
MSR-VTT & Find the clip that corresponds to the described scene in the given video. & Recall@1 \\
DiDeMo & Find a video that includes the following described scenes. & Recall@1 \\
CRB-G & Find the video according to the general text description. & Recall@1 \\
CRB-S & Find the video according to the spatial description. & Recall@1 \\
VDC-O & Find the video according to the object description. & Recall@1 \\
CRB-T & Find the video according to the temporal description. & Recall@1 \\
CMRB & Find the video according to the camera motion description. & Recall@10 \\
DREAM-E & Find the video according to the text description. & Recall@1 \\
LoVR-TH & Find the video according to text description about video theme information. & Recall@10 \\
PEV-K & Find the video according to the text description of a series of keywords. & Recall@1 \\
LoVR-V & Find the long video according to the long text description. & Recall@1 \\
VDC-D & Find the video according to the detailed text description. & Recall@1 \\
MS-TI & Find the video clip that corresponds to the given text and the given image. & Precision@1 \\
MS-TV & Find the video clip that corresponds to the given text and the given video. & Precision@1 \\
MSRVTT-I2V & Find the video according to the image. & Recall@1 \\
LoVR-C2V & Find the original long video according to the short video clip. & Recall@1 \\
\bottomrule
\end{tabular}
\end{table}

\subsection{Evaluation Pipeline.}\label{app:evaluation_pipeline}
Our evaluation is built atop MTEB~\cite{muennighoff2022mteb}. It decouples the evaluation engine from model architecture—supporting everything from sentence transformers to custom multimodal encoders—via a standardized interface. The pipeline operates in three phases: (1) dynamic task orchestration, (2) configurable model execution, and (3) generation of metrics and diagnostics.
For instruction-sensitive tasks, it dynamically injects domain-specific prompts to mirror real-world conditions for actionable diagnostics.
Custom benchmarks are integrated seamlessly through dedicated loaders.
In this framework, expensive tasks are deferred for efficiency, model initialization (precision, multi-GPU) is abstracted via a factory, and data loading supports both online and offline modes for robustness across environments.
Engineering safeguards, including explicit multi-GPU management, clean shutdowns, and offline, first retry logic—ensure reliability at scale.

\subsection{Training Data}\label{app:training_data}

To build a robust and versatile multimodal retrieval system, our models are trained on a large-scale, diverse mixture of text, image, and video data. All tasks are uniformly formulated as instruction-guided retrieval, a strategy designed to foster a unified representation space that can adeptly handle a wide array of queries. The training data is organized into two primary collections: a main mixture of widely-used public datasets (Table~\ref{tab:main_training_data}), and a synthesized set from our Universal Video Retrieval Dataset (UVRD) suite (Table~\ref{tab:uvrd_data}).

We follow existing works~\cite{li2023towards,Zhang_2025_CVPR} to construct text-only~(e.g., MSMARCO) and image-centric data~(e.g., CIRR).
In addition, we prepare video datasets using the following strategies.
\begin{itemize}
    \item \textbf{VAST}~\cite{chen2023vast}: We randomly select one sentence from the multiple vision captions as the text query for each video.
    \item \textbf{InternVid-FLT}~\cite{wang2023internvid}: We drop about $300$K low-quality videos and use the left text-video pairs.
    \item \textbf{PE-Video}~\cite{bolya2025perception}: We choose the $\text{human\_caption}$ refined on $\text{model\_caption}$ as the textual description of videos.
    \item \textbf{WebVid}~\cite{bain2021frozen}: We refuse the queries captioning over one video to impede the generality of text, which excludes $5$M videos in final training approximately.
\end{itemize}

\begin{table}[ht!]
\centering
\begin{threeparttable}
\caption{Configuration of the main training data mixture for our 3B and 7B models. `K` denotes thousands, `M` denotes millions.}
\label{tab:main_training_data}
\footnotesize
\begin{tabularx}{\linewidth}{X c r r c c}
\toprule
\textbf{Dataset} & \textbf{Task}\tnote{a} & \multicolumn{2}{c}{\textbf{Sample Size}} & \textbf{Neg.}\tnote{b} & \textbf{BS} \\
\cmidrule(lr){3-4}
& & \textbf{3B} & \textbf{7B} & & \\
\midrule
\multicolumn{6}{l}{\textbf{Part 1: Text-only Data}} \\
MSMARCO & T→T & 300K & 500K & 2 & 64 \\
HotpotQA & T→T & \multicolumn{2}{c}{69K} & 1 & 64 \\
WebQA & T→T & \multicolumn{2}{c}{11K} & 1 & 32 \\
\midrule
\multicolumn{6}{l}{\textbf{Part 2: Image-centric Data}} \\
CIRR & TI→T & \multicolumn{2}{c}{16K} & 1 & 32 \\
Fashion200K & T→I & \multicolumn{2}{c}{4K} & 1 & 32 \\
Nights & I→I & \multicolumn{2}{c}{13K} & 1 & 32 \\
OVEN & TI→TI & 20K & 30K & 1 & 64 \\
OVEN & TI→T & 20K & 30K & 1 & 64 \\
VisualNews & T→I & 40K & 60K & 1 & 64 \\
EDIS & T→TI & \multicolumn{2}{c}{12K} & 1 & 32 \\
FashionIQ & TI→I & \multicolumn{2}{c}{4K} & 1 & 32 \\
MSCOCO & I→T & \multicolumn{2}{c}{4K} & 1 & 32 \\
REMUQ & TI→T & \multicolumn{2}{c}{5K} & 1 & 32 \\
WebQA & T→TI & \multicolumn{2}{c}{12K} & 1 & 32 \\
LLAVA & TI→T & 20K & 30K & 1 & 64 \\
EVQA & TI→TI & 20K & 30K & 1 & 64 \\
CC3M & T→I & 200K & 300K & 3 & 32 \\
CC3M & I→T & 100K & 200K & 2 & 64 \\
Laion & T→I & 300K & 500K & 3 & 32 \\
Laion & I→T & \multicolumn{2}{c}{200K} & 2 & 64 \\
ImageNet & I→T & 100K & 200K & 2 & 32 \\
VL3-Syn7M (short) & T→I & 300K & 500K & 3 & 32 \\
VL3-Syn7M (short) & I→T & 100K & 200K & 2 & 64 \\
VL3-Syn7M (detailed) & T→I & 200K & 300K & 3 & 32 \\
VL3-Syn7M (detailed) & I→T & 100K & 200K & 2 & 64 \\
VISTA & TI→I & \multicolumn{2}{c}{100K} & 2 & 32 \\
VISTA & T→TI & \multicolumn{2}{c}{20K} & 1 & 64 \\
\midrule
\multicolumn{6}{l}{\textbf{Part 3: Video-centric Data}} \\
VAST & T→V & \multicolumn{2}{c}{1.6M} & 0 & 32 \\
InternVid-FLT & T→V & \multicolumn{2}{c}{1.7M} & 0 & 32 \\
PE-Video & T→V & \multicolumn{2}{c}{104K} & 1 & 64 \\
WebVid & T→V & \multicolumn{2}{c}{5.4M} & 0 & 32 \\
\bottomrule
\end{tabularx}
\begin{tablenotes}
    \item[a] \textbf{Task}: T=Text, I=Image, V=Video. The format is Query→Corpus.
    \item[b] \textbf{Neg.}: Number of explicit hard negatives per positive. '0' indicates use of in-batch negatives only.
\end{tablenotes}
\end{threeparttable}
\end{table}

\begin{table}[ht!]
\centering
\begin{threeparttable}
\caption{Configuration of UVRD. Sample sizes are identical for both 3B and 7B models.}
\label{tab:uvrd_data}
\footnotesize
\begin{tabularx}{\linewidth}{X c c c c}
\toprule
\textbf{Dataset} & \textbf{Task}\tnote{a} & \textbf{Sample Size} & \textbf{Neg.}\tnote{b} & \textbf{BS} \\
\midrule
UVRD-T2T & T→T & 100K & 0 & 64 \\
UVRD-T2I & T→I & 210K & 0 & 32 \\
UVRD-T2V & T→V & 879K & 1 & 64 \\
UVRD-TI2V & TI→V & 89K & 1 & 64 \\
UVRD-TV2V & TV→V & 35K & 1 & 64 \\
UVRD-I2V & I→V & 200K & 0 & 64 \\
UVRD-V2V & V→V & 36K & 0 & 64 \\
\bottomrule
\end{tabularx}
\begin{tablenotes}
    \item[a] \textbf{Task}: T=Text, I=Image, V=Video. The format is Query→Corpus.
    \item[b] \textbf{Neg.}: Number of explicit hard negatives per positive. '0' indicates use of in-batch negatives only.
\end{tablenotes}
\end{threeparttable}
\end{table}

As the aforementioned methodology, we train our model based on contrastive learning with pre-mined explicit hard negatives and in-batch negatives.
To maintain a balanced training diet and manage computational load, we employ a sophisticated data sampling strategy. For extremely large datasets, we sample a fixed number of instances. To amplify the learning signal from certain high-value or complex datasets, we may apply an upsampling strategy by repeating their data.

The scale of our training data is substantial. For our 3B parameter model, we prepared $12.55$ million instances. To further leverage the capacity of our 7B parameter model, we increase the sampling rate for several large-scale datasets, bringing the total instances to $13.73$ million. A detailed breakdown of the data composition is provided in Table~\ref{tab:data_summary}.

\begin{table}[ht!]
\centering
\caption{Total number of instances for our 3B and 7B models across all data categories.}
\label{tab:data_summary}
\begin{tabularx}{\linewidth}{X r r}
\toprule
\textbf{Data Category} & \multicolumn{2}{c}{\textbf{Total Sample Size}} \\
\cmidrule(lr){2-3}
& \textbf{3B Model} & \textbf{7B Model} \\
\midrule
Collected: Text-only & 380K & 580K \\
Collected: Image-centric & 1.82M & 2.80M \\
Collected: Video-centric & 8.80M & 8.80M \\
Synthesized: UVRD (All Modalities) & 1.55M & 1.55M \\
\midrule[1.5pt]
\bfseries Total & \bfseries 12.55M & \bfseries 13.73M \\
\bottomrule
\end{tabularx}
\end{table}

\subsection{Dataset Construction Pipeline Details}\label{app:data_construct}

Current multimodal retrieval research is hindered by fragmented, ad-hoc dataset construction. We introduce a unified, object-oriented framework that elevates this process to a rigorous science, balancing conceptual clarity with engineering robustness for scalable, reproducible benchmarking.

The framework is anchored in a canonical tripartite abstraction: every retrieval task is decomposed into a \textsc{Corpus}, a set of \textsc{Queries}, and a \textsc{Relevance Mapping}. This schema, enforced by an abstract base class, is a conceptual invariant that ensures structural consistency across all datasets, from text-to-video to complex composed queries, enabling seamless model and evaluation compatibility.

The expressiveness of this data construction pipeline stems from polymorphic specialization via inheritance, encapsulated in three core strategies:

\paragraph{Modality as Configuration.}
Modality is a dynamic parameter. By overriding a single method, a dataset effortlessly transitions between modalities (text, image, video, or composite), transforming benchmarks like MSRVTT into image-to-video tasks with minimal code.

\paragraph{Task Derivation via Inheritance.}
Complex tasks are composed of simpler ones. A text+image-to-video task inherits its base structure and extends the query schema. Variants are derived by overriding specific data-access methods, turning benchmark creation into a modular, hypothesis-driven workflow.

\paragraph{Engineering for Scale and Integrity.}
Scalability and data quality are first-class principles. Large datasets (e.g., InternVid-FLT) leverage chunked, parallel processing. Crucially, proactive curation is embedded: automated validators check video integrity (resolution, frame count), while repair mechanisms fix common errors using \texttt{ffmpeg}, ensuring failures reflect retrieval challenges, i.e, not data corruption.

\subsection{Prompt for Synthetic Video Retrieval Data Generation}\label{app:prompt}

To ensure high-quality, diverse, and controllable synthetic video retrieval data generation at scale, we design and deploy a suite of structured prompts to instruct MLLMs for captioning. Our pipeline operates in four parts:

\begin{enumerate}
    \item \textbf{Raw~(Or Weakly Annotated) Video (Re-)~Captioning}: Enhance raw or uncaptioned videos with rich, diverse textual descriptions.
    \item \textbf{Text-Image Composed Retrieval}: Generate queries that combine reference images with video content for fine-grained retrieval.
    \item \textbf{Text-Video Composed Retrieval}: Generate queries that combine short reference clips with target videos for practical temporal or perspective-based retrieval.
    \item \textbf{Frame Image Captioning}: Annotate individual video frames with dynamic-aware captions for auxiliary training signals.
\end{enumerate}

All prompts enforce strict output formatting, factual grounding, and stylistic diversity to ensure dataset quality and coverage. \texttt{\{raw\_caption\}} represents an optional, potentially low-quality human-provided or auto-generated caption associated with the video. Placeholders (e.g., \texttt{\{readability\}} and \texttt{\{education\_level\}}) are dynamically instantiated during batch generation.

Besides, we can generate text-to-text retrieval data from the multiple video/frame captions by randomly matching any two sentences in the caption list.

\begin{promptbox}{Synthetic Video Captioning Prompt}
\begin{lstlisting}[style=promptstyle]
Generate 5 distinct and high-quality ENGLISH captions for the provided video. Please first visually understand the video file and analyze the video frame-by-frame in depth before captioning.

The original video caption is {raw_caption}. (Ignore if empty.)

Each caption must be:
  1. The final answer MUST only be a JSON dict of captions where the key is the caption number and the value is a single-paragraph caption.
  2. Factually accurate and descriptive - include only what is clearly visible or reasonably inferable from the video.
  3. Focus exclusively on visible content; do not mention absences or speculate about unseen elements.

Content Requirements (per caption):
  1. Spatial Details (30-60%): Describe location, setting, key environmental elements, objects, and their spatial relationships. Include notable visual features.
  2. Temporal Flow (30-60%): Capture event sequence and action dynamics - movements, interactions, transitions - as they unfold over time.
  3. Theme/Background/Style/Meaning/Highlight/Camera (0-20%): Describe observable emotional tone, narrative style, thematic elements, highlighting frames, or significant camera movements/angles.
  4. Others (0-10%).

Key Guidelines:
  1. Do not invent fictional elements, dialogue, or backstory not visible in the video.
  2. Use the following varied sentence styles (one per caption):
       - concise and punchy summary,
       - spatial-temporal richly descriptive,
       - abstract understanding,
       - keywords-only,
       - partially relevant information.
  3. Ensure diversity: avoid repetition in wording, focus, or rhythm; at least one caption must be <20 words and one >100 words; balance objectivity with vivid sensory language; randomly omit minor details in 1 or 2 captions.
  4. Ensure readability matches {readability} and is appropriate for {education_level} readers.

Now generate these captions in strict JSON format:
{
   "1": <caption text 1: concise and punchy summary, 10-25 words>,
   "2": <caption text 2: spatial-temporal richly descriptive, 80-200 words>,
   "3": <caption text 3: abstract understanding, 30-100 words>,
   "4": <caption text 4: keywords-only, 5-30 words>,
   "5": <caption text 5: partially relevant information, 10-70 words>
}
\end{lstlisting}
\end{promptbox}

\begin{promptbox}{Synthetic Text-Image Composed Video Retrieval Prompt}
\begin{lstlisting}[style=promptstyle]
You are an information retrieval expert specializing in high-value text-image composed queries. Your sole objective is to generate queries that significantly enhance video retrieval performance by effectively combining a reference image and video content.

## What Makes a HIGH-QUALITY Query (Non-Negotiable)
A truly valuable query must satisfy ALL of these criteria:

1. COMBINATION NECESSITY (Most Critical)
   - The query MUST become meaningless or significantly less specific if either the image or video is removed.
   - Example (HIGH-QUALITY): "the person FROM REFERENCE IMAGE wearing red jacket now skiing"
   - Example (LOW-QUALITY): "a person skiing" (works without image)

2. SEMANTIC PRECISION
   - Must accurately reflect BOTH the visual content of the reference image AND the video.
   - Must reference at least ONE specific visual attribute from the image (not generic descriptions).
   - Example (HIGH-QUALITY): "matching the blue hat FROM PHOTO, now running through park"
   - Example (LOW-QUALITY): "someone similar to image moving" (too vague)

3. RETRIEVAL EFFECTIVENESS
   - Must narrow search results by at least 50\% compared to text-only queries.
   - Must contain actionable constraints that differentiate from 90\% of videos in the database.
   - Example (HIGH-QUALITY): "the woman WITH PONYTAIL FROM IMAGE entering building at 2PM"
   - Example (LOW-QUALITY): "a woman walking" (too broad)

4. LOGICAL COHERENCE
   - Must maintain subject-verb-object consistency across modalities.
   - Must avoid semantic contradictions between image attributes and video actions.
   - Example (HIGH-QUALITY): "the dog FROM REFERENCE PHOTO chasing a ball"
   - Example (LOW-QUALITY): "the red jacket FROM IMAGE running down hill" (jackets don't run)

5. PRACTICAL UTILITY
   - Must solve a real-world ambiguity that neither text nor image could resolve alone.
   - Must reflect how actual users would express their information need.
   - Example (HIGH-QUALITY): "same person AS IN PHOTO but wearing blue instead of red"
   - Example (LOW-QUALITY): "the image shows a person and the video shows action" (no real combination)

## High-Value Query Generation Framework
Follow this structured approach:

1. DEEP ANALYSIS PHASE (Mandatory)
   a) Reference Image Analysis:
      - Identify 1-4 SPECIFIC visual attributes (e.g., "red jacket", "ponytail", "blue hat")
      - Determine primary subject with discriminative features
      - Note what CANNOT be determined from image (e.g., action, scene)
   
   b) Video Content Analysis:
      - Identify primary action using precise verbs (e.g., "running", "entering", "chasing")
      - Note scene context and temporal elements
      - Determine what CHANGES from the static image reference

2. COMBINATION STRATEGY SELECTION
   Choose ONE primary strategy:

   A) IDENTITY PRESERVATION + ACTION CHANGE
      - Structure: [Binding phrase] + [Image attribute] + [Video action]
      - Example: "the man FROM REFERENCE IMAGE in blue shirt now running"
   
   B) IDENTITY PRESERVATION + SCENE MIGRATION
      - Structure: [Binding phrase] + [Image attribute] + [Scene transition]
      - Example: "same person AS IN PHOTO moving from office to park"
   
   C) IDENTITY PRESERVATION + NEGATIVE CONSTRAINT
      - Structure: [Binding phrase] + [Negative constraint] + [Video state]
      - Example: "not wearing red jacket FROM IMAGE but blue, walking"
   
   D) RELATIONAL TRANSFER
      - Structure: [Binding phrase] + [Relationship description]
      - Example: "the dog FROM REFERENCE PHOTO chasing a ball"

3. QUALITY ENHANCEMENT TECHNIQUES
   - Binding Precision: Use "FROM REFERENCE IMAGE", not "like the picture"
   - Attribute Specificity: Use concrete features ("red jacket", not "clothing")
   - Action Verbs: Use present continuous tense ("running", not "runs")
   - Context Enrichment: Add 1 relevant scene descriptor ("in park", "near building")
   - Noise Handling: For low-quality inputs, use "resembling" but maintain specificity
   - Audience Adaptation: Ensure readability matches {readability} and suits {education_level} readers.

4. MANDATORY QUALITY CHECK
   Before finalizing, verify ALL:
   - [ ] Explicit binding to reference image
   - [ ] References SPECIFIC visual attribute from image
   - [ ] Describes DYNAMIC ELEMENT not in static image
   - [ ] Loses specificity if image removed
   - [ ] Contains actionable retrieval constraint
   - [ ] Maintains logical subject-action consistency
   - [ ] Solves real-world ambiguity

## What to AVOID
- Generic descriptions ignoring image specificity
- Redundant mentions of obvious image content
- Semantic contradictions (e.g., "the jacket is running")
- Overly precise details not visible in inputs
- Standalone video descriptions without image binding
- Vague terms: "something", "thing", "area"
- Excessive length without added value (>50 words)

## Output Format
- Output ONLY a JSON object with one key: "query"
- Value must be the generated query sentence
- No additional text or formatting

Generate the query for:
Reference Image: <image_input>
Video Clip: <video_input>
\end{lstlisting}
\end{promptbox}

\begin{promptbox}{Synthetic Text-Video Composed Video Retrieval Prompt}
\begin{lstlisting}[style=promptstyle]
You are an information retrieval expert specializing in practical text-video composed queries. Your objective is to generate queries that effectively combine a short reference clip with a target video for real-world retrieval.

## Practical Context Understanding
In real-world scenarios:
- Reference is typically a SHORT CLIP (2-10 seconds)
- Clip is usually HIGHLY RELEVANT to target video (same source or similar content)
- Common relationships: temporal continuation, perspective variation, quality differences, minor action variations
- Query should reflect how users actually search

## What Makes a HIGH-QUALITY Query (Practical Focus)
Must satisfy these criteria:

1. USEFUL COMBINATION (Most Important)
   - Leverages reference clip to specify what text alone cannot
   - Example (HIGH-QUALITY): "the same person FROM REFERENCE CLIP continuing to run after the jump"
   - Example (LOW-QUALITY): "a person running" (ignores reference)

2. PRACTICAL PRECISION
   - References at least ONE observable feature from reference clip
   - Example (HIGH-QUALITY): "matching the red jacket FROM REFERENCE, now entering building"
   - Example (LOW-QUALITY): "someone similar to clip moving" (too vague)

3. REAL-WORLD UTILITY
   - Helps find videos difficult to retrieve with text alone
   - Example (HIGH-QUALITY): "same action AS IN REFERENCE but from front angle"
   - Example (LOW-QUALITY): "the clip shows action" (no retrieval value)

4. NATURAL EXPRESSION
   - Sounds like how a real user would phrase it
   - Example (HIGH-QUALITY): "what happens right after this moment?"
   - Example (LOW-QUALITY): "temporal continuation of the current visual sequence" (too academic)

## Practical Query Generation Framework

1. REFERENCE CLIP ANALYSIS
   - Identify 1-3 KEY OBSERVABLE FEATURES (e.g., "red jacket", "starting pose", "mid-action")
   - Determine most distinctive visual or action element
   - Note what is CLEARLY VISIBLE (avoid guessing)

2. TARGET VIDEO RELATIONSHIP ASSESSMENT
   Determine relationship type:
   
   A) TEMPORAL CONTINUATION
      - Reference is earlier part of same sequence
      - Query focus: "what happens next" or "continuing action"
   
   B) PERSPECTIVE VARIATION
      - Same action from different angle/view
      - Query focus: "same action from different angle"
   
   C) QUALITY/CONDITION VARIATION
      - Same action with different lighting/resolution
      - Query focus: "same scene in better lighting"
   
   D) MINOR ACTION VARIATION
      - Slightly different execution of similar action
      - Query focus: "same person but running instead of walking"

3. QUERY CONSTRUCTION
   - Start with binding phrase: "FROM REFERENCE CLIP", "AS IN REFERENCE", etc.
   - Reference 1-2 specific observable features from clip
   - Describe relationship to target video clearly
   - Keep natural and practical (5-15 words typically)
   - For temporal: focus on "what happens next"
   - For perspective: specify desired viewpoint
   - For quality: specify desired condition
   - For action: specify the change

4. PRACTICAL QUALITY CHECK
   Ask before finalizing:
   - Would this help me find what I'm looking for?
   - Does it add value beyond describing the target?
   - Would a real user phrase it this way?
   - Is everything mentioned clearly visible in reference?

## What to AVOID
- Overly academic or technical language
- References to features NOT clearly visible
- Excessive precision about timing ("exactly 3.2 seconds later")
- Queries that work equally well without reference
- Generic descriptions: "similar video", "related content"
- Making unsupported assumptions

## Output Format
- Output ONLY a JSON object with key: "query"
- Value must be the generated query string
- No additional text, explanations, or formatting

Generate the query for:
Reference Clip: <reference_clip_input>
Target Video: <target_video_input>
\end{lstlisting}
\end{promptbox}

\begin{promptbox}{Synthetic Frame Captioning Prompt}
\begin{lstlisting}[style=promptstyle]
Generate 5 distinct and high-quality ENGLISH captions for the provided image (a frame extracted from a video) based solely on visual content. Please first visually understand the image and analyze it in depth before captioning.

Each caption must be:
  1. The final answer MUST only be a JSON dict of captions where the key is the caption number and the value is a single-paragraph caption.
  2. Factually accurate and descriptive - include only what is clearly visible; captions should also be consistent with the short-term video context.
  3. Focus exclusively on visible content; do not mention absences or speculate about unseen context.

Content Requirements (per caption):
  1. Spatial Details (50-70%): Describe location, setting, key objects, spatial relationships, colors, lighting, composition. Include instantaneous action states derived from visible posture/motion cues.
  2. Temporal Snapshots (0-30%): Describe the frozen moment's temporal state (movements, interactions, transitions) ONLY if visually provable. Avoid implying sequence, duration, or speed.
  3. Theme/Style/Composition (10-20%): Cover emotional tone, camera angle, lighting style, or artistic elements directly observable.
  4. Others (0-10%).

Key Guidelines:
  1. Do not invent fictional elements or backstory. If action is ambiguous, describe neutrally. Never use future/past tense or speculative phrases.
  2. Use varied sentence styles (randomly assign one per caption):
       - concise spatial-temporal snapshot,
       - spatially rich descriptive,
       - abstract spatial interpretation,
       - keywords with temporal anchors,
       - minimalist spatial focus.
  3. Ensure diversity: avoid repetition in wording, focus, or rhythm; at least one caption <20 words and one >100 words; balance objectivity with vivid sensory language; randomly omit minor details in 1-2 captions.
  4. Ensure readability matches {readability} and is appropriate for {education_level} readers.

Now generate these captions in strict JSON format:
{
   "1": <caption text 1: concise and punchy summary, 10-25 words>,
   "2": <caption text 2: spatial-temporal richly descriptive, 50-200 words>,
   "3": <caption text 3: abstract understanding, 30-100 words>,
   "4": <caption text 4: keywords-only, 5-30 words>,
   "5": <caption text 5: partially relevant information, 10-70 words>
}
\end{lstlisting}
\end{promptbox}

\paragraph{Prompt Utilization.}
These prompts are designed for a scalable, schema-constrained generation pipeline for synthesizing diverse video-centric annotations and queries. The system unifies four tasks under a single modular framework.
Each task is governed by a structured prompt template with dynamic control slots (e.g., readability, education level), ensuring linguistic and semantic diversity. Input modalities (image, video, or both) are automatically routed and embedded via a distributed multimodal LLM (Keye-VL-8B~\cite{team2025kwai}, 32K context), with outputs rigorously validated against JSON schemas for structural correctness.
The pipeline supports sharded, resumable batch generation with quality-aware sampling, enabling the production of millions of grounded, stylistically varied synthetic instances.

\subsection{Model Architecture Details}
\label{app:arch}

The GVE model is architecturally derived from Qwen2.5-VL~\cite{bai2025qwen25vl}, repurposed as a fixed-length multimodal encoder by removing its autoregressive head. Its core function is to map arbitrarily composed inputs—text, image, or video—into a shared $d$-dimensional embedding space, preserving cross-modal alignment inherited from pretraining.

Input fusion begins with tokenization and visual encoding. Text is converted into a sequence of token IDs $\mathbf{X}_t \in \mathbb{Z}^{B \times T_t}$, while images and video frames are encoded into visual token sequences $\mathbf{X}_v \in \mathbb{R}^{B \times T_v \times d}$, where $T_v = \text{THW} / p_{s}^2$ for images, and $T_v = K \cdot (\text{THW} / p_{s}^2)$ for videos with $K$ uniformly sampled frames. Note that frame embeddings will be added with absolute time encoding. Critically, $K$ must satisfy $K \bmod p_{t} = 0$ (default: $p_{t} = p_{s} = 2$) to maintain alignment with the vision encoder’s 3D spatiotemporal grid. The processor then injects these visual tokens into the textual sequence by replacing placeholder tokens (\texttt{<image>}, \texttt{<video>}), producing a fused input $\mathbf{X}_{\text{fused}} \in \mathbb{R}^{B \times T \times d}$, where $T = T_t + \sum T_v$. This scatter-based fusion preserves positional coherence and enables interleaved modality composition, which is essential for compositional query understanding in multimodal embeddings.
The final embedding $\mathbf{e}^{(i)} \in \mathbb{R}^d$ for the $i$-th instance is extracted from the last attended token in the sequence:
\begin{equation*}
    \mathbf{e}^{(i)} = \frac{\mathbf{h}^{(i)}_{p_i}}{\|\mathbf{h}^{(i)}_{p_i}\|_2}, \quad \text{where } p_i = \max \{ j \mid \mathbf{M}^{(i)}_j = 1 \},
\end{equation*}
and $\mathbf{M}^{(i)}$ is the attention mask for instance $i$. This position corresponds to the EOS token in left-padded sequences or the final non-pad token in right-padded ones. The choice is motivated by the observation that in instruction-tuned MLLMs, the final token often encapsulates the model’s response intent, making it semantically aligned with the user’s retrieval goal.

\subsection{Training Implementation and Hyperparameters}\label{app:hyperparameters}

\paragraph{Parameter-Efficient Tuning.}
To facilitate parameter-efficient fine-tuning (PEFT), we employ Low-Rank Adaptation (LoRA)~\cite{hu2021lora}. Our strategy involves a targeted application of LoRA to the language components of the model, specifically the \texttt{q\_proj, v\_proj, k\_proj, up\_proj, down\_proj,} and \texttt{gate\_proj} modules. Crucially, the entire visual backbone and the base token embedding layer are kept frozen. This approach focuses adaptation on high-level semantic and cross-modal reasoning while preserving the powerful pretrained visual features. We enable FlashAttention-2~\cite{dao2023flashattention} to accelerate training and reduce memory. The final embedding for each input is obtained via last-token pooling on the last hidden layer, followed by L2 normalization to project embeddings onto the unit hypersphere for stable cosine similarity computations.

\paragraph{Optimizer and Training Dynamics.}
The model is trained using the AdamW~\cite{loshchilov2017decoupled} optimizer with a learning rate of \(3 \times 10^{-5}\) and a weight decay of $0.1$. A cosine learning rate scheduler is used. Training is performed in BFloat16 (\texttt{bf16}) mixed-precision. The entire training process is managed under the DeepSpeed framework. For video inputs, we uniformly sample $8$ frames per clip at a rate of $1.0$ FPS. By default, we use $32$ NVIDIA A100 GPUs, each with 80GB of memory. Therefore, the overall batch size is at least $1024$.

\paragraph{Memory Optimization and Distributed Strategy.}
To manage GPU memory, we enable gradient checkpointing, which recomputes intermediate activations during the backward pass. While our framework supports more advanced techniques like Gradient Cache, it was disabled in favor of this standard approach. To leverage our multi-GPU setup, we enable cross-device negative sharing. This strategy gathers embeddings from all GPUs, effectively multiplying the pool of in-batch negatives by the number of devices. This enriches the negative set for the contrastive loss computation on each GPU, leading to a stronger training signal without increasing per-device memory load.

\paragraph{Contrastive Learning and Stability.}
The core of our training is an InfoNCE-style contrastive loss with a temperature of $0.03$. The use of cross-device negatives starts from the first step. We also enhance our training logs with contrastive-specific metrics, including the average scores of positive pairs and the average margin between positive and hard-negative pairs, providing crucial insights into the model's learning dynamics.

\paragraph{Hyperparameter Summary.}
A comprehensive summary of all key hyperparameters is provided in Table \ref{tab:training_hyperparameters}.

\begin{table}[ht!]
\centering
\footnotesize
\caption{Key hyperparameters used for model training.}
\label{tab:training_hyperparameters}
\begin{tabular}{ll}
\toprule
\textbf{Parameter} & \textbf{Value} \\
\midrule
\multicolumn{2}{l}{\textbf{Model \& LoRA Architecture}} \\
Base Model & \texttt{Qwen2.5-VL-3B-Instruct} or \texttt{Qwen2.5-VL-7B-Instruct} \\
PEFT Method & LoRA \\
LoRA Rank (\(r\)) & 16 \\
LoRA Alpha (\(\alpha\)) & 32 \\
LoRA Dropout & 0.1 \\
LoRA Target Modules & \texttt{q\_proj, v\_proj, k\_proj, up\_proj, down\_proj, gate\_proj} \\
Frozen Components & Visual Backbone, Token Embeddings \\
\midrule
\multicolumn{2}{l}{\textbf{Optimizer \& Training}} \\
Optimizer & AdamW \\
Learning Rate & \(3 \times 10^{-5}\) \\
Weight Decay & 0.1 \\
LR Scheduler & Cosine \\
Training Epochs & 3 \\
Precision & BF16 \\
Gradient Checkpointing & Enabled \\
Seed & 42 \\
\midrule
\multicolumn{2}{l}{\textbf{Contrastive Learning}} \\
Temperature & 0.03 \\
In-batch Negatives & Enabled \\
Cross-Device Negatives & Enabled \\
\midrule
\multicolumn{2}{l}{\textbf{Data \& Preprocessing}} \\
Video Frames per Clip & 8 \\
Video Sampling Rate & 1.0 FPS \\
Dataloader Workers & 1 \\
\bottomrule
\end{tabular}
\end{table}

\subsection{Baseline Details}\label{app:baseline_details}

Here we present more details of baseline models tested on our benchmark in Table~\ref{tab:models}, including full model names, abbreviations, architectures, sizes, and training data types. Based on these properties, we analyze to discover potential performance knowledge and dependencies in our experimental part.

\begin{table}[ht!]
\centering
\footnotesize
\setlength{\tabcolsep}{2pt}
\caption{Model Abbreviations, Architectures, Parameter Sizes, and Training Data Types. The checkmark (\checkmark) indicates the model was trained on the corresponding data pair types.}
\label{tab:models}
\begin{tabular}{@{}lllc ccc@{}} 
\toprule
& & & & \multicolumn{3}{c}{\textbf{Contrastive Training Data Pairs}} \\
\cmidrule(lr){5-7}
\textbf{Full Model Name} & \textbf{Abbreviation} & \textbf{Architecture} & \textbf{Size} & \textbf{Text-Text} & \textbf{Text-Image} & \textbf{Text-Video} \\
\midrule
CLIP4Clip           & CLIP4Clip     & CLIP-based & 87M    & - & - & \checkmark \\
ViCLIP              & ViCLIP        & CLIP-based & 0.4B   & - & - & \checkmark \\
VideoCLIP-XL        & VideoCLIP-XL  & CLIP-based & 0.4B   & - & - & \checkmark \\
LanguageBind-Video-Huge-V1.5 & LanguageBind & CLIP-based & 1.2B & - & - & \checkmark \\
InternVideo2-Stage2-1B & InternVideo2-1B & CLIP-based & 1.4B & - & - & \checkmark \\
InternVideo2-Stage2-6B & InternVideo2-6B & CLIP-based & 6.4B & - & - & \checkmark \\
\midrule
gme-Qwen2-VL-2B-Instruct & GME-2B   & MLLM-based & 2.2B & \checkmark & \checkmark & - \\
Unite-Base-Qwen2-VL-2B   & Unite-2B & MLLM-based & 2.2B & \checkmark & \checkmark & \checkmark \\
VLM2Vec-V2.0             & VLM2Vec-V2 & MLLM-based & 2.2B & \checkmark & \checkmark & \checkmark \\
BGE-VL-v1.5-mmeb         & BGE-VL     & MLLM-based & 7.6B & \checkmark & \checkmark & - \\
UniME-LLaVA-OneVision-7B & UniME-7B  & MLLM-based & 8.0B & \checkmark & \checkmark & - \\
B3-Qwen2-7B              & B3-7B & MLLM-based & 8.3B & \checkmark & \checkmark & - \\
gme-Qwen2-VL-7B-Instruct & GME-7B    & MLLM-based & 8.3B & \checkmark & \checkmark & - \\
Unite-Base-Qwen2-VL-7B   & Unite-7B  & MLLM-based & 8.3B & \checkmark & \checkmark & \checkmark \\
\midrule
GVE-3B          & GVE-3B    & MLLM-based & 3.8B & \checkmark & \checkmark & \checkmark \\
GVE-7B          & GVE-7B    & MLLM-based & 8.3B & \checkmark & \checkmark & \checkmark \\
\bottomrule
\end{tabular}
\end{table}

\subsection{Training Dynamics}\label{app:training_dynamics}

\begin{figure}[ht!]
    \centering
    \includegraphics[width=\linewidth]{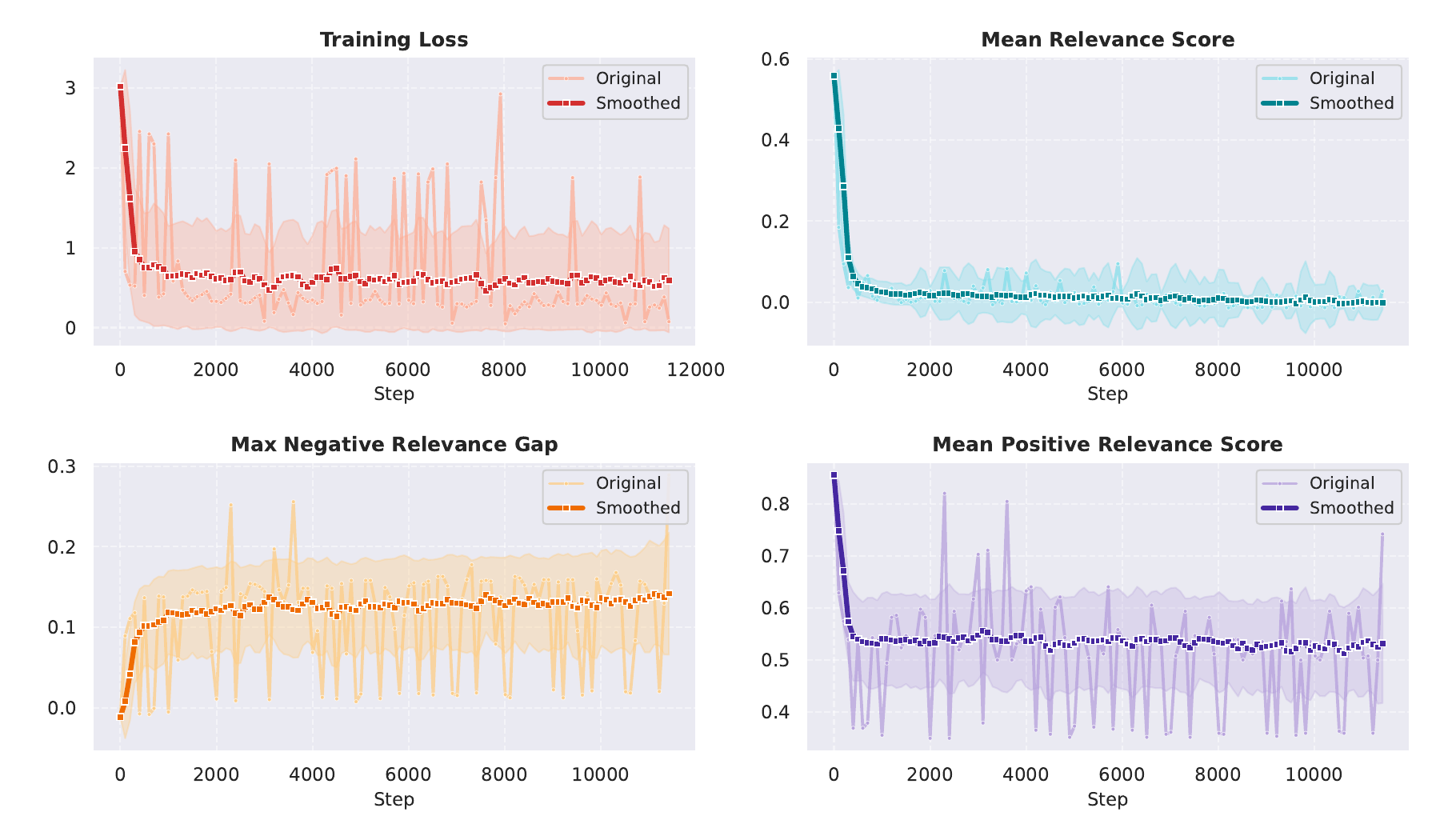}
    \caption{Training dynamics of \texttt{GVE-3B} across four metrics.}
    \label{fig:train_3b}
\end{figure}

\begin{figure}[ht!]
    \centering
    \includegraphics[width=\linewidth]{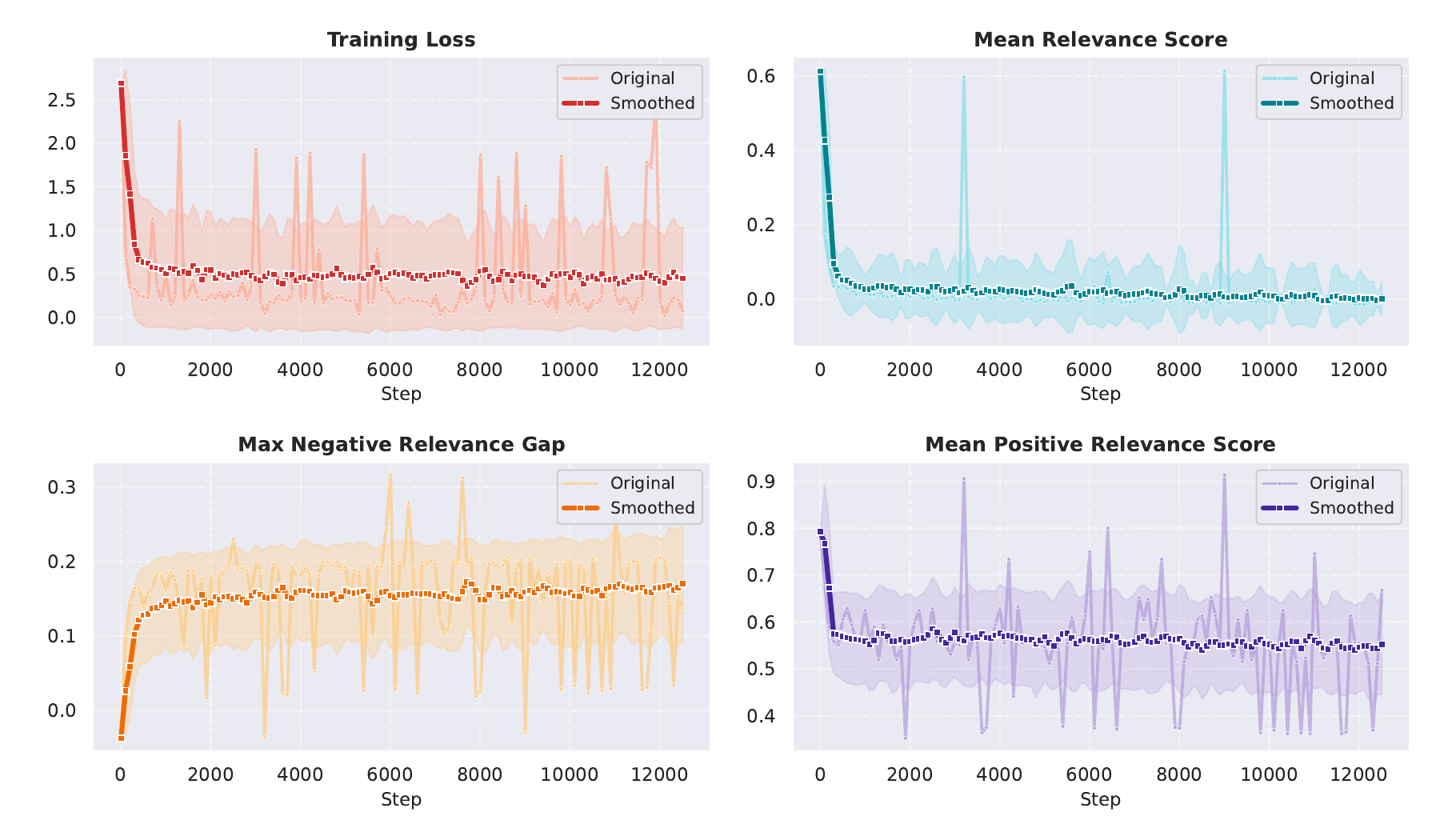}
    \caption{Training dynamics of \texttt{GVE-7B} across four metrics.}
    \label{fig:train_7b}
\end{figure}

We monitor four key metrics during training: \textit{Training Loss}, \textit{Mean Score}, \textit{Max Negative Gap}, and \textit{Mean Positive Score} in Figure~\ref{fig:train_3b} and Figure~\ref{fig:train_7b}. To ensure robust visualization, we mitigate outlier effects, followed by $200$-step moving average smoothing. Original trajectories (subsampled every $100$ steps) are plotted alongside smoothed trends, with ±1 standard deviation bands indicating local volatility.

\subsection{Experiments of Training-time Scaling: More Results of Data Scaling}\label{app:data_scaling}

\begin{figure}[ht!]
    \centering
    \includegraphics[width=\linewidth]{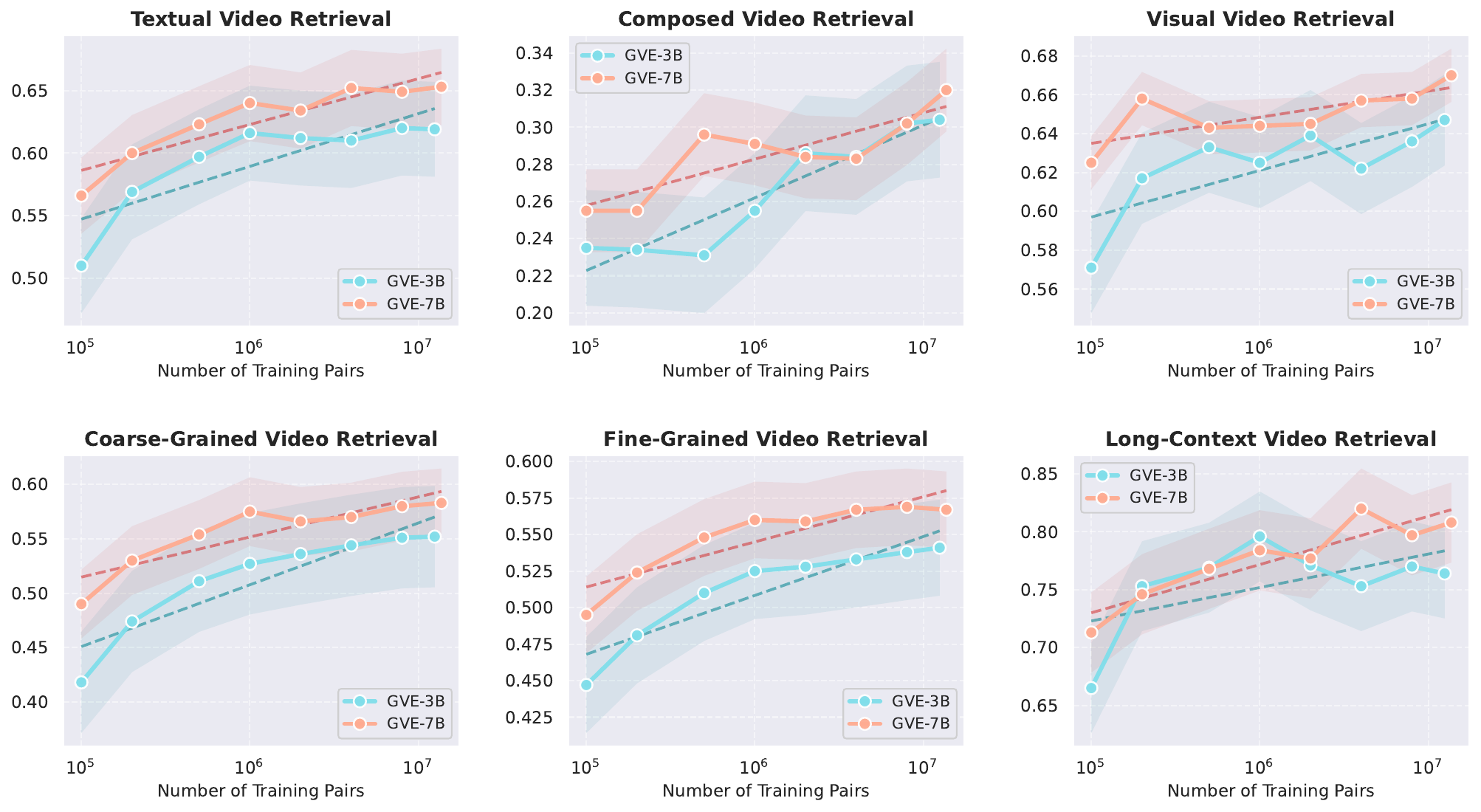}
    \caption{Performance effect from data scaling for GVE-3B and GVE-7B in detail.}
    \label{fig:scaling_detail}
\end{figure}

Along with Section~\ref{sec:data_scaling}, we depict more experiments for six abilities in data scaling for \texttt{GVE-3B} and \texttt{GVE-7B} in Figure~\ref{fig:scaling_detail}. Shaded bands show $\pm1$ std; dashed curves are log-linear fits for visual guidance. X-axis is log-scaled to reflect scaling law dynamics.

We fit a logarithmic scaling law $y = a \ln x + b$ to quantify gains per decade of data (10$\times$ increase). GVE-3B consistently shows higher relative gains: in compositional retrieval (CMP), it improves by $+0.039$ ($14.7\%$), nearly double GVE-7B’s $+0.025$ ($8.7\%$); in coarse-grained tasks (CG), it gains $+0.057$ ($11.1\%$), far exceeding GVE-7B’s $+0.037$ ($6.6\%$). Textual (TXT) and fine-grained (FG) retrieval follow similar trends, with GVE-3B improving by $+0.042$ ($7.1\%$) and $+0.040$ ($7.9\%$), respectively, versus GVE-7B’s $+0.037$ ($5.8\%$) and $+0.031$ ($5.6\%$). Visual retrieval (VIS) scales weakly for both ($+0.024$/$3.8\%$ for GVE-3B, $+0.013$/$2.1\%$ for GVE-7B). Notably, only in long-context retrieval (LC) does GVE-7B outperform GVE-3B in both absolute ($+0.042$ vs. $+0.029$) and relative gain ($5.4\%$ vs. $3.8\%$).

This reveals a task-dependent scaling trade-off: smaller models (GVE-3B) scale more efficiently in semantic and compositional tasks, while larger models (GVE-7B) uniquely excel in long-context modeling — suggesting that model size should be chosen not only for capacity, but for alignment with the data-scaling profile of the target task.

\subsection{Experiments of Test-time Scaling}\label{app:scaling}

We investigate the impact of scaling two test-time parameters: the number of sampled frames and the maximum tokens per frame (i.e., effective resolution). We analyze each parameter in isolation: when scaling the frame count from $8$ to $48$, the maximum token count is fixed at $200$; conversely, when scaling the token count from $200$ to $800$, the frame count is fixed at $8$.

\paragraph{Temporal Scaling: the Number of Sampled Frames.}

\begin{figure}[ht!]
    \centering
    \includegraphics[width=\linewidth]{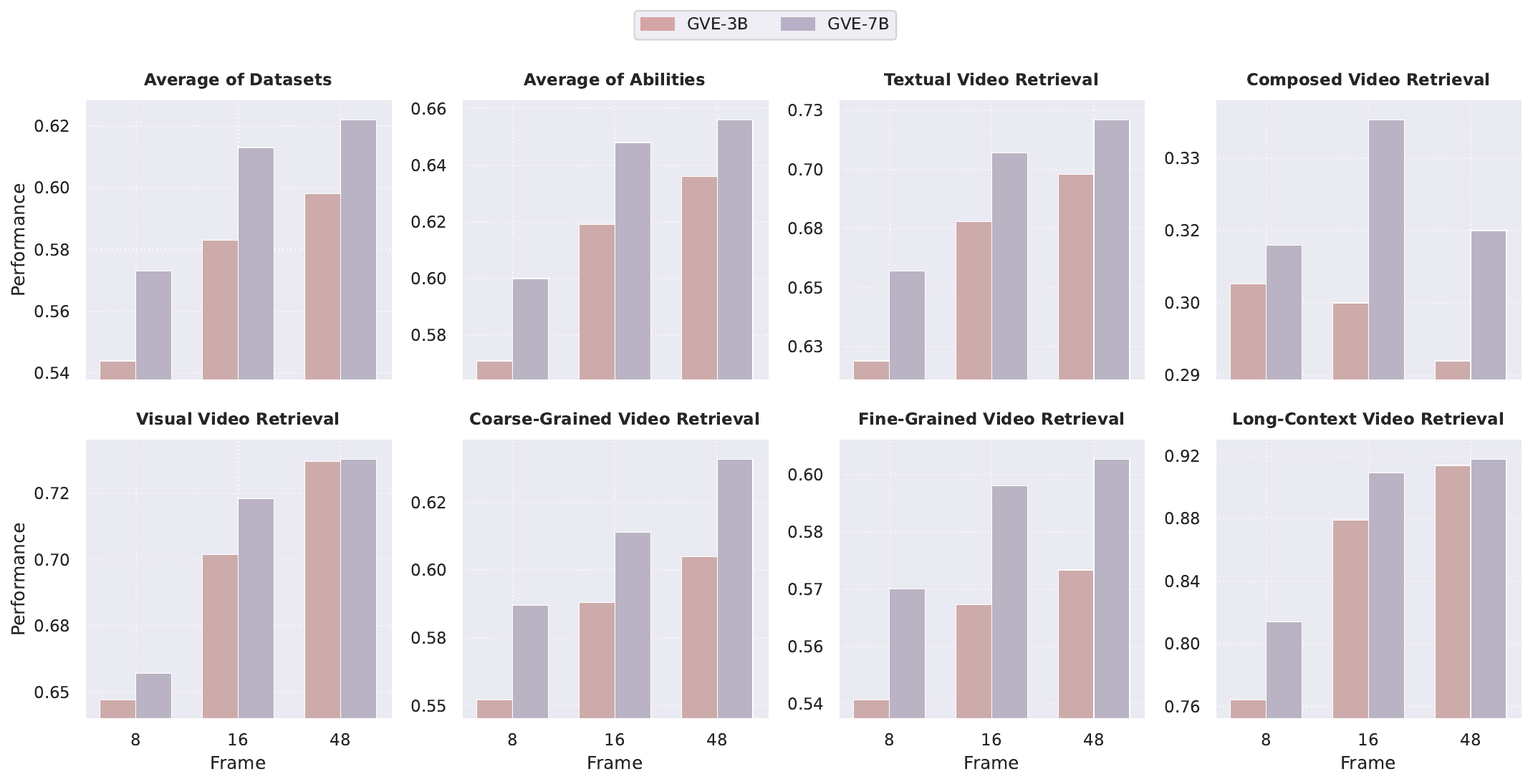}
    \caption{Model performance as a function of the number of sampled frames at inference time (max tokens fixed at $200$).}
    \label{fig:scaling_frame}
\end{figure}

As shown in Figure~\ref{fig:scaling_frame}, increasing the number of sampled frames generally improves performance. The most substantial gains appear in the Long-Context (LC) task (GVE-3B: +$19.6\%$, GVE-7B: +$12.8\%$), underscoring the value of denser temporal sampling for long-range reasoning. However, the Compositional (CMP) task shows a slight performance degradation, suggesting its sensitivity to potentially redundant visual cues from excessive frames. Furthermore, while GVE-7B consistently outperforms GVE-3B, the performance gap narrows as frames increase (from $0.029$ to $0.024$ on AVG-D), indicating that the larger model is more efficient at extracting information from sparser inputs.

\paragraph{Spatial Scaling: the Maximum Tokens Per Frame.}

\begin{figure}[ht!]
    \centering
    \includegraphics[width=\linewidth]{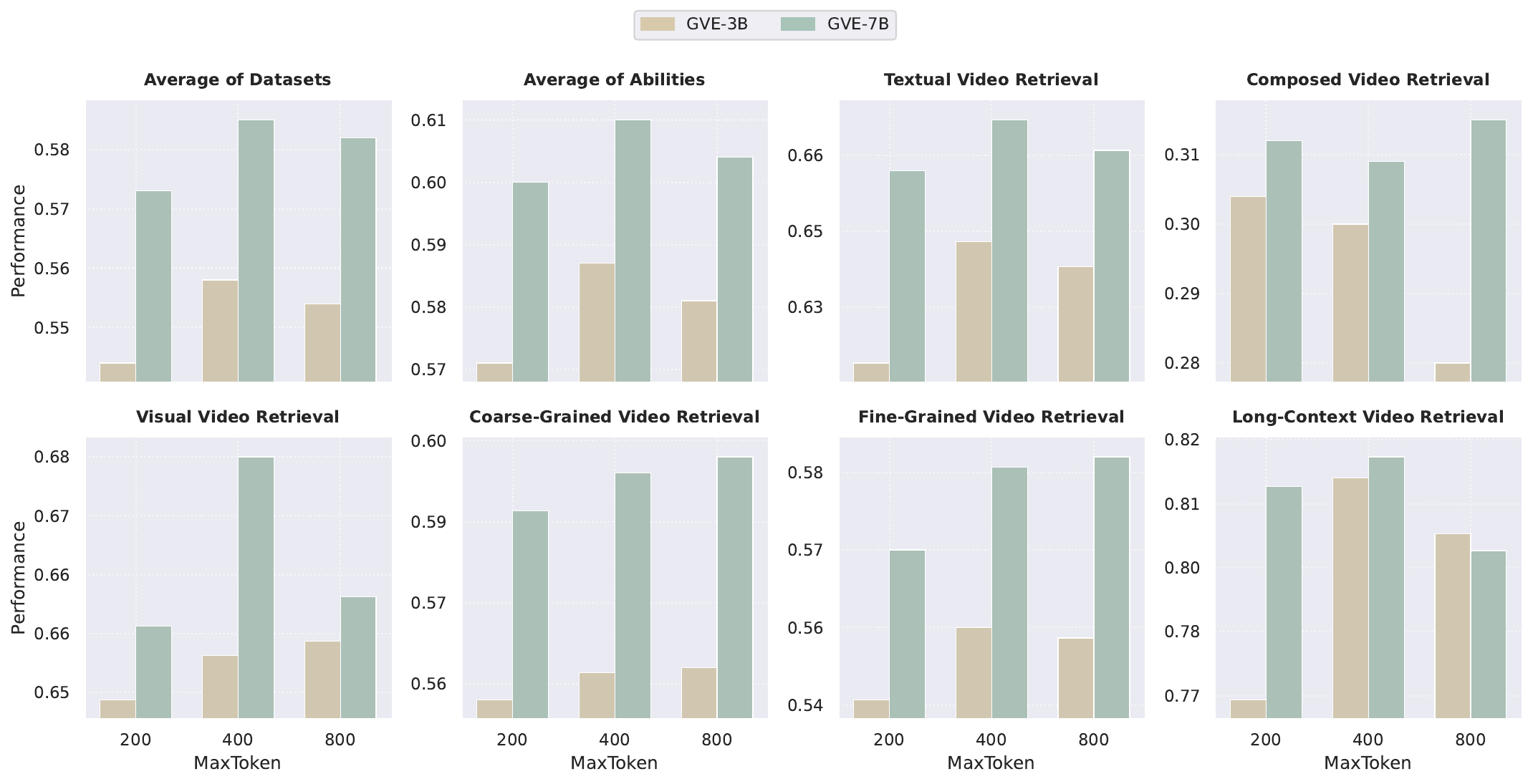}
    \caption{Model performance versus the maximum number of tokens per frame (frame count fixed at $8$), determining the resolution of frames in resizing.}
    \label{fig:scaling_resolution}
\end{figure}

In contrast, increasing the token budget yields non-monotonic returns (Figure~\ref{fig:scaling_resolution}). Performance for most tasks peaks around $400$ tokens and then declines. For instance, in the LC task, both models' scores drop when moving from $400$ to $800$ tokens. This suggests that for long-context tasks, excessive per-frame detail can dilute attention from salient features rather than aid understanding. The CMP task's performance again deteriorates with more tokens, reinforcing its sensitivity to input redundancy.

\paragraph{Key Findings and Implications.}

Our scaling analysis reveals four key findings:
\begin{itemize}
    \item \textbf{Temporal Scaling (More Frames):} Provides a reliable, though diminishing, performance boost, especially for long-range tasks.
    \item \textbf{Spatial Scaling (More Tokens):} Exhibits a clear optimal point (approx. $400$ tokens). Exceeding it degrades performance, indicating a trade-off between detail and distraction.
    \item \textbf{Model Scale Efficiency:} The larger model (GVE-7B) benefits less from input scaling, suggesting a greater intrinsic capacity to process sparse inputs effectively.
\end{itemize}

For test-time enhancement, temporal scaling is a robust strategy, whereas naive spatial scaling is not. This distinction underscores a critical insight: performance is not merely a function of total input data, but of its effective composition. It strongly motivates a shift from brute-force data scaling towards adaptive input mechanisms, such as dynamic token or frame selection, that can intelligently manage information density.

\subsection{Experiments of Video Classification}\label{app:video_classification}

We test the video embeddings for video classification by following the MMEB-V2~\cite{meng2025vlm2vec}.
Note that we do not train on these datasets for zero-shot evaluations, while other baselines may include them in optimization.
Specifically, we evaluate video embedding models on five diverse classification benchmarks, including \textbf{Kinetics-700~(K700)}~\cite{carreira2019short}, \textbf{UCF101}~\cite{soomro2012ucf101}, \textbf{HMDB51}~\cite{kuehne2011hmdb}, \textbf{SomethingSomething-V2~(SSV2)}~\cite{goyal2017something}, and \textbf{Breakfast}~\cite{kuehne2014language}, covering fine-grained interactions, open-domain actions, and procedural activities. Each task is cast as video-to-text retrieval over the full label set, using standardized validation/test splits.
The results are provided in Table~\ref{tab:video_classification}.
Our results show that \texttt{LanguageBind} achieves the highest mean accuracy ($0.553$), followed closely by \texttt{GVE-7B} ($0.526$) and \texttt{InternVideo2-6B} ($0.526$), demonstrating the efficacy of unified multimodal representations. However, performance remains suboptimal on temporally complex datasets such as Breakfast (best: $0.453$) and fine-grained SSV2 (best: $0.569$), revealing a critical bottleneck in modeling dynamic physical interactions.

\begin{table}[ht!]
\centering
\footnotesize
\caption{Video classification accuracy comparison across different models and datasets. For each column: highest score is \textbf{bolded}, second-highest is \underline{underlined}.}
\label{tab:video_classification}
\begin{tabular}{l| >{\columncolor{gray!15}}c | ccccc}
\toprule
\textbf{Model} & \textbf{AVG} & \textbf{K700} & \textbf{UCF101} & \textbf{HMDB51} & \textbf{SSV2} & \textbf{Breakfast} \\
\midrule
CLIP4Clip       & 0.378 & 0.395 & 0.596 & 0.413 & 0.308 & 0.176 \\
InternVideo2-6B & \textbf{0.526} & \textbf{0.554} & 0.572 & 0.480 & \textbf{0.569} & \textbf{0.453} \\
VLM2Vec-V2      & 0.393 & 0.380 & 0.600 & 0.409 & 0.428 & 0.148 \\
GME-7B          & 0.374 & 0.397 & 0.547 & 0.479 & 0.306 & 0.143 \\
UniME-7B        & 0.306 & 0.388 & 0.377 & 0.407 & 0.190 & 0.166 \\
Unite-7B        & \underline{0.519} & 0.537 & \underline{0.752} & \textbf{0.534} & 0.513 & 0.261 \\
\midrule
GVE-3B          & 0.476 & 0.489 & 0.661 & 0.483 & 0.471 & 0.277 \\
GVE-7B          & \textbf{0.526} & \underline{0.540} & \textbf{0.757} & \underline{0.525} & \underline{0.521} & \underline{0.289} \\
\bottomrule
\end{tabular}
\end{table}

\subsection{Limitations}\label{app:limitations}
Our study has several practical limitations that stem from scope boundaries and experimental design.

First, all evaluations are conducted in a vision-only setting, excluding audio, transcripts, or metadata. While this aligns with standard practice in video-text retrieval, it limits applicability to real-world scenarios where multimodal cues (e.g., sound) are essential for disambiguation.

Second, inference uses a fixed protocol: videos are uniformly sampled into exactly $8$ frames, and input sequences are truncated at $8192$ tokens. This may disadvantage tasks requiring adaptive frame selection (e.g., sparse event detection) or longer context (e.g., hour-long videos).

Third, although UVRB covers $16$ datasets across diverse tasks, it does not include specialized domains such as medical, industrial, or surveillance videos, where visual semantics and query intent differ significantly from general-domain content.

Finally, training GVE, especially the 7B variant, requires substantial computational resources, limiting accessibility for researchers with a constrained infrastructure. Efficient variants and training strategies are left for future work.

\end{document}